\documentclass[letterpaper]{article} % DO NOT CHANGE THIS
\usepackage{aaai2027}  % DO NOT CHANGE THIS
\usepackage{multirow}
\nocopyright

\usepackage[hyphens]{url}  % DO NOT CHANGE THIS
\usepackage{graphicx} % DO NOT CHANGE THIS
\usepackage{natbib}  % DO NOT CHANGE THIS AND DO NOT ADD ANY OPTIONS TO IT
\usepackage{caption} % DO NOT CHANGE THIS AND DO NOT ADD ANY OPTIONS TO IT
\usepackage{algorithm}
\usepackage{algorithmic}
\usepackage{xcolor}
\usepackage{amsmath}
\usepackage{amssymb}
\usepackage{newfloat}
\usepackage{listings}
\DeclareCaptionStyle{ruled}{labelfont=normalfont,labelsep=colon,strut=off} % DO NOT CHANGE THIS
\floatstyle{ruled}
\newfloat{listing}{tb}{lst}{}
\floatname{listing}{Listing}

\usepackage{booktabs}

\newcommand{\systemname}{ReTouch}

\newcommand{\Encoder}{Tactile-Patch Encoder}
\title{

ReTouch: Empowering Contact-Rich Dexterous Manipulation with \\ Online-Refined Tactile Prediction

}
\author{
    Shiqi Zhang\textsuperscript{\rm 1}\equalcontrib,
    Xin Zhang\textsuperscript{\rm 1}\equalcontrib,
    Yedong Shen\textsuperscript{\rm 1},
    Yao Li\textsuperscript{\rm 1}\corresponding, 
    Yuxuan Gao\textsuperscript{\rm 1},
    Sha Zhang\textsuperscript{\rm 3},
    Yuan Zhang\textsuperscript{\rm 2},
    Kaixue Long\textsuperscript{\rm 1},
    Jiajia Wu\textsuperscript{\rm 2},
    Jia Pan\textsuperscript{\rm 2},
    Jiajun Deng\textsuperscript{\rm 1},
    Yanyong Zhang\textsuperscript{\rm 1}\corresponding
}

\affiliations{
    \textsuperscript{\rm 1}University of Science and Technology of China\\
    \textsuperscript{\rm 2}iFLYTEK\\
    \textsuperscript{\rm 3}The Chinese University of Hong Kong
}
\begin{document}

\maketitle

\begin{abstract}
Fusing tactile signals has proven effective for contact-rich manipulation, enabling robots to perceive contact states and adapt to rapidly changing physical interactions. Yet effectively integrating tactile feedback into dexterous manipulation remains underexplored. In this work, we introduce ReTouch, a vision-language-action model (VLA) that supports contact-rich dexterous manipulation through tactile predictions continually refined online using execution-time feedback. ReTouch builds on two main innovations for tactile representation and closed-loop action generation. First, its \Encoder\ represents tactile observations as structured tactile patch features that preserve finger identity and local contact structure, providing contact cues for fine-grained dexterous control. Second, its high-frequency action module jointly predicts future tactile states and action chunks, and refines both using incoming tactile feedback during execution. This closed-loop refinement keeps tactile predictions aligned with evolving physical interactions, enabling responsive action correction and improving robustness to contact changes and execution errors. We further introduce XHT-Dataset, comprising 900 real-world demonstrations across seven contact-rich tasks collected on an XHand–UR7e platform, and evaluate ReTouch through closed-loop real-robot experiments. ReTouch surpasses the strongest baseline by 18.4 and 23.8 percentage points in average success rate under standard and challenging conditions, respectively, demonstrating its effectiveness and robustness.
\end{abstract}

\section{Introduction}
\begin{figure*}[t]
    \centering
    \includegraphics[width=\textwidth]{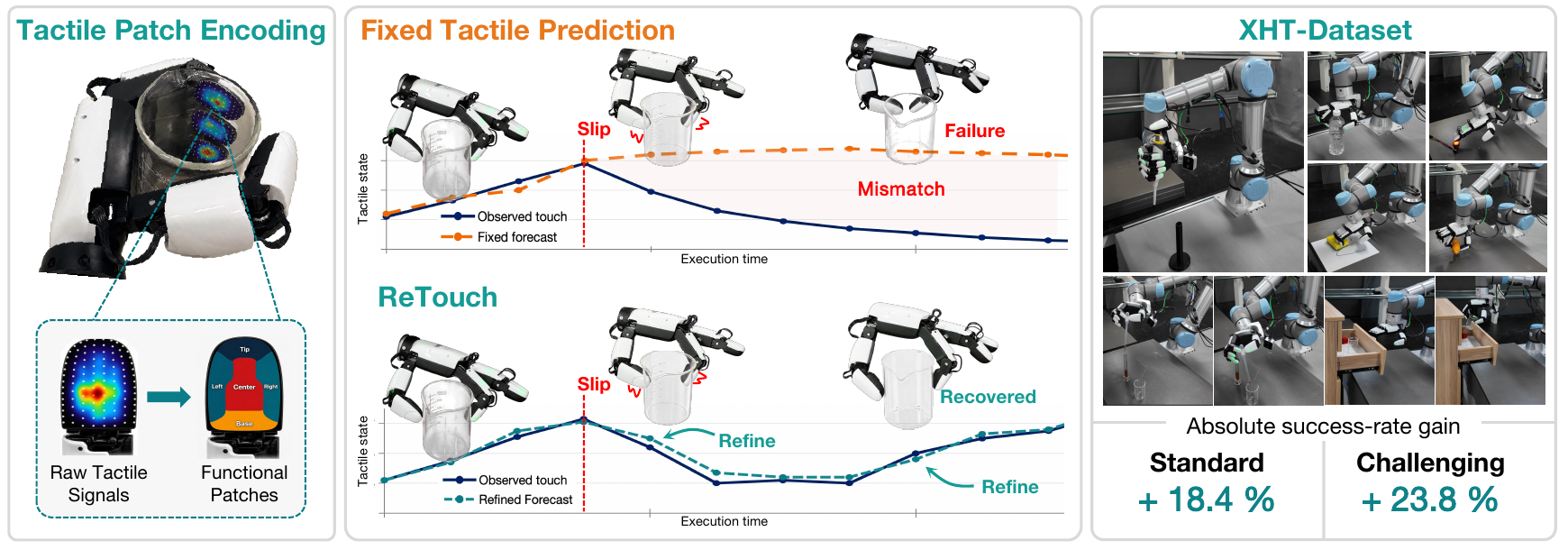}
    \caption{
Overview of \systemname. 
Left: the Tactile-Patch Encoder organizes raw dexterous-hand tactile signals into structured tactile patch features. 
Middle: fixed tactile forecasts can diverge from the actual contact state after slippage, while \systemname\ continually refines tactile predictions using the latest tactile feedback and updates action chunks accordingly. 
Right: XHT-Dataset: seven real-world contact-rich tasks on XHand--UR7e under standard and challenging settings.
}
\label{fig:overview}
\end{figure*}

% 我们的贡献总结如下：
% 我们提出 \systemname，一种Empowering Dexterous Manipulation with Online-Refined Tactile Prediction的VLA框架。在执行过程中，\systemname 利用最新触觉反馈，online修正触觉预测，从而修正动作。

% 我们提出一种面向灵巧手的结构化触觉表征学习方法。\Encoder 根据手指身份和指尖接触拓扑组织密集触觉信号。在此基础上，双动作专家在动作生成监督下学习与动作相关的未来触觉潜变量。

% 我们搭建了XHand--UR7e触觉操作平台，并构建了XHT，其中包含七项接触密集型任务的900条真实世界示范。\systemname 在标准测试和分布外测试中的平均成功率分别比最强基线高出18.4 points和23.8 points；消融实验验证了其关键组件的有效性。
%  【 灵巧操作重要 之前有人做了dexVLA 】

Contact-rich dexterous manipulation requires robots to continuously track local hand--object interactions, including which fingers are in contact, whether the object is slipping, and whether multi-finger contact remains stable. Vision-Language-Action models (VLAs)~\cite{black2024pi_0,intelligence2025pi_,kim2024openvla} provide instruction understanding and scene-level perception, but vision alone cannot reliably reveal contact locations, forces, or grasp stability~\cite{calandra2017feeling,calandra2018more}. This perceptual gap is especially pronounced in multi-finger manipulation, where hand--object occlusions hide contact regions; tactile sensing therefore provides direct local feedback for fine-grained action adjustment~\cite{heng2026vitacformerlearningcrossmodalrepresentation}.

Recent tactile-aware policies have expanded the role of touch beyond passive observation.
Direct-fusion methods incorporate current tactile signals as an additional modality to improve contact awareness~\cite{huang2025tactile,zhang2026tacvlacontactawaretactilefusion,huang2026tafvlatactileforcealignmentvisionlanguageaction}.
Reactive methods use high-frequency tactile or force feedback to revise actions online in response to slippage, contact shifts, or force variations~\cite{xue2025reactive,li2026favla,niu2026t}.
% More recently, predictive methods forecast future tactile states or contact forces to guide action generation~\cite{zang2026tacforesight}, while predictive-reactive methods combine such forecasts with online action refinement~\cite{zhou2026touchworld,niu2026learning}.
% More recently, predictive methods forecast future tactile states or contact forces to support action generation or representation learning~\cite{zang2026tacforesight,niu2026learning}, while predictive-reactive methods combine such forecasts with online action refinement~\cite{zhou2026touchworld}.
Recently, predictive methods forecast future tactile states or contact forces to support action generation~\cite{heng2026vitacformerlearningcrossmodalrepresentation,zang2026tacforesight,niu2026learning,lou2026dreamtacunifiedtactileworld}, while predictive-reactive methods combine such forecasts with online action refinement~\cite{he2026fawamforceawareworldaction,zheng2026omnivtavisuotactileworldmodeling,zhang2026unitacvlaunifiedtactileunderstanding,zhou2026touchworld}.
Together, these developments extend touch from passive observation to reactive contact feedback and predictive action guidance.

% 尽管取得上述进展，现有触觉感知策略应用于灵巧操作仍面临两个关键挑战。第一，灵巧手触觉信号维度高，而接触响应具有局部稀疏性。现有方法通常将触觉信号直接展平为通用触觉token~\cite{huang2025tactile}，或将整根手指的触觉信号联合编码为单一表征~\cite{niu2026t}。这些表示未显式保留逐指局部接触拓扑，因此需要一种按手指身份和局部接触区域组织信号的结构化触觉表征，以支持精细的接触调整。
% 第二，多指动作高度耦合，会产生复杂且快速变化的接触动态，使未来触觉状态难以准确预测。现有预测式方法通常在动作块执行前生成一次未来触觉预测，并在执行期间保持该预测固定~\cite{zang2026tacforesight,lou2026dreamtacunifiedtactileworld}。接触滑移或执行误差可能使其迅速偏离真实交互状态。预测—反应式方法虽然能够利用最新触觉反馈修正动作~\cite{zhou2026touchworld,niu2026learning}，但用于指导动作的触觉预测并未同步更新，从而造成预测与控制之间的不匹配，并可能误导后续动作更新。因此，未来触觉预测不应在动作块执行期间保持固定，而应根据最新触觉反馈持续在线修正。

Despite this progress, applying tactile-aware policies to dexterous manipulation presents two key challenges. First, dexterous hands 
produce dense tactile observations, while task-relevant contacts are typically sparse and localized to specific fingers and fingertip regions.
% Existing methods typically flatten tactile signals into generic tactile tokens~\cite{huang2025tactile} or encode tactile signals from each finger into a single representation~\cite{niu2026t}. 
% Existing methods typically flatten tactile signals into generic tactile tokens~\cite{huang2025tactile} or compress each finger's tactile signals without explicitly modeling its local contact regions~\cite{niu2026t}.
% These representations do not explicitly preserve finger-wise local contact topology.
Existing methods typically flatten tactile signals into generic tactile tokens~\cite{huang2025tactile} or compress each finger's temporal force history into representations~\cite{niu2026t}.
Such representations obscure the finger-wise spatial structure of contact, making it difficult for the policy to associate local contact changes with fine-grained finger adjustments. 
% A structured tactile representation is therefore needed to explicitly preserve both finger identity and local contact topology. % Existing predictive methods typically generate a future tactile prediction before an action chunk and keep it fixed throughout execution~\cite{zang2026tacforesight,lou2026dreamtacunifiedtactileworld}. 
% 【让预测信号敏感 不是实际敏感】
Second, contact states in multi-finger manipulation can change rapidly because of slippage, contact displacement, and execution errors, causing future tactile predictions generated at the beginning of an action chunk to become stale during execution. Existing methods typically do not continually update these predictions using the latest tactile feedback within the action chunk ~\cite{zang2026tacforesight,lou2026dreamtacunifiedtactileworld},
% Contact slippage or execution errors may then cause the prediction to diverge rapidly from the actual interaction state.
% Although predictive-reactive methods use the latest tactile feedback to refine actions~\cite{zhou2026touchworld,niu2026learning}, the tactile prediction guiding those actions is not updated accordingly. This mismatch between prediction and control may misguide subsequent action updates, as illustrated in Fig.~\ref{fig:overview}. Future tactile prediction should therefore be refined online with the latest tactile feedback, rather than remain a fixed prior during execution.
while predictive-reactive methods use the latest tactile feedback to refine actions~\cite{zhou2026touchworld}. Meanwhile, the tactile prediction guiding action is not re-estimated within each high-rate tactile update.
Consequently, the predicted tactile state can become stale and misguide subsequent action refinement (Fig.~\ref{fig:overview}).

To address these challenges, we introduce \systemname, a VLA framework for dexterous manipulation with online-refined tactile prediction, as shown in Fig.~\ref{fig:overview}.
\systemname\ first uses a \Encoder~to organize tactile signals from dexterous hands. The encoder divides the taxels on each finger into five patches according to functional contact regions and adds finger-identity and patch-position embeddings. The resulting structured tactile patch features allow the model to associate local contact responses with specific fingers and contact regions, supporting fine-grained fingertip adjustments.
To ensure that future tactile predictions capture action-relevant information, we introduce a Hindsight Action Expert (HAE) during training. The HAE uses ground-truth future tactile patch features to produce action-relevant target latents, while a Foresight Action Expert (FAE) learns to predict the corresponding future tactile latents from current observations.
% During execution, the FAE is \systemname's high-frequency action module and runs asynchronously with the vision-language model (VLM). At the beginning of each action chunk, the VLM encodes and caches the visual-language context, while the FAE uses the latest tactile feedback to continually refine future tactile predictions and update the action chunk accordingly. 
During execution, the FAE serves as ReTouch's high-frequency action module and runs asynchronously with the VLM. Within each action chunk, the FAE continually uses observed tactile feedback to refine future tactile predictions, keeping them aligned with the ongoing physical interaction and improving their accuracy, thereby providing more reliable guidance for action-chunk updates.
To evaluate \systemname, we build XHT-Dataset on a real-world XHand--UR7e platform, with 900 demonstrations across seven contact-rich dexterous manipulation tasks. \systemname\ achieves average success-rate gains of +18.4\% and +23.8\% over the strongest baseline under standard and challenging settings, respectively.

Our contributions are summarized as follows:

% \begin{itemize}
%     \item 我们提出了一种在线触觉预测修正机制。在执行过程中，高频动作模块利用最新的触觉反馈，联合更新未来触觉预测和动作块，使触觉预测持续与不断变化的物理交互保持一致，并随着反馈更新逐步提高预测准确性，从而为高频动作修正提供更可靠的指导。

%     \item 我们设计了一个 \Encoder，将多指触觉观测组织为保留手指身份和局部接触结构的触觉块特征，使模型能够将局部接触响应与对应的手指和接触区域相关联，从而支持精细的指尖动作调整。

%     \item 我们在 XHand--UR7e 真实机器人平台上构建了 XHT-Dataset。该数据集由900条真实世界示范组成，覆盖七项接触密集型灵巧操作任务。我们在标准设置和四种挑战性设置下通过闭环实验验证了 ReTouch 的有效性和鲁棒性。
% \end{itemize}

% \begin{itemize}
%     \item We propose \systemname, a VLA that asynchronously refines future tactile predictions and action chunks using the latest tactile feedback during execution.

%     \item We introduce a \Encoder~that organizes dexterous-hand tactile signals into structured tactile patch features, explicitly encoding finger identity, local contact region, and local tactile response.

%     % \item We build XHT-Dataset with seven contact-rich dexterous manipulation tasks and validate the effectiveness and robustness of \systemname\ under standard and four challenging settings.
%     \item We build XHT-Dataset, comprising seven contact-rich dexterous manipulation tasks, and systematically validate the effectiveness and robustness of \systemname\ across standard and four challenging settings.

% \end{itemize}

\begin{itemize}
    \item We introduce an online tactile-prediction refinement mechanism that jointly updates future tactile predictions and action chunks using the latest feedback, improving prediction accuracy as interactions evolve and providing reliable guidance for high-frequency action refinement.

    \item We design a \Encoder\ for dexterous hands that preserves finger identity and local contact structure, allowing the model to associate local contacts with specific finger regions for fine-grained adjustments.

    \item We build XHT-Dataset with seven contact-rich dexterous manipulation tasks on a XHand--UR7e platform, and show the effectiveness and robustness of ReTouch through experiments under standard and challenging settings.
\end{itemize}

% To address these challenges, we introduce \systemname, a VLA framework for dexterous manipulation with online-refined tactile prediction. \systemname\ first uses a \Encoder~to encode tactile signals from dexterous hands. The encoder groups the taxels on each finger into five functional contact patches and augments each patch with finger-identity and patch-position embeddings. The resulting structured tactile patch features jointly encode finger identity, local contact region, and local tactile response. 
% Based on these features, \systemname\ introduces two action experts to learn action-relevant future tactile latents under action-generation supervision. The training-only Hindsight Action Expert (HAE) extracts target latents from ground-truth future tactile patch features, whereas the Foresight Action Expert (FAE) predicts the corresponding future tactile latents from currently available observations and aligns them with these targets. 
% At execution time, \systemname\ removes the HAE and retains the FAE as its high-frequency action module. Within each action chunk, the vision-language model (VLM) encodes and caches the visual-language context once, whereas the FAE is invoked asynchronously at a higher frequency. At each invocation, the FAE uses the latest tactile feedback to refine future tactile latents and update the action chunk accordingly. An asymmetric causal mask prevents action-token information from leaking into tactile predictions while allowing the updated future tactile latents to directly guide action generation.

% !TEX root = ../AnonymousSubmission2027.tex
\section{Related Work}

Tactile and force sensing are increasingly used to ground
learning-based manipulation in local physical interactions.
Recent tactile-aware VLAs integrate contact observations
through token-level fusion, contact-aware gating, semantic
alignment, or tactile--force alignment
\cite{huang2025tactile,
zhang2026tacvlacontactawaretactilefusion,
cheng2026omnivtlavisiontactilelanguageactionmodelssemanticaligned,
huang2026tafvlatactileforcealignmentvisionlanguageaction}.
Complementary methods preserve geometric structure through
shared 3D visual--tactile representations or hand-centric
kinematic anchoring
\cite{huang20253dvitaclearningfinegrainedmanipulation,
huang2026spatiallyanchoredtactileawareness}.
These works establish several mechanisms for
grounding action prediction in physical contact, but largely
condition policies on current or recent observations.

A second line of work exploits the higher bandwidth of
tactile and force feedback for closed-loop control. RDP
decouples low-rate trajectory generation from high-rate
tactile refinement; FAVLA uses predicted force variation to
schedule a force-conditioned action expert; and T-Rex
extends asynchronous tactile refinement to dexterous
manipulation through a variable-rate MoT and temporal
tactile encoding
\cite{xue2025reactive,li2026favla,niu2026t}.
In parallel, predictive methods model how contact may evolve.
ViTacFormer predicts future tactile observations for
cross-modal representation learning, TacForeSight conditions
action generation on predicted tactile latents, and Dream-Tac,
Tactile-WAM, and VT-WAM jointly model future tactile states with robot actions
\cite{heng2026vitacformerlearningcrossmodalrepresentation,
zang2026tacforesight,
lou2026dreamtacunifiedtactileworld,
wu2026tactilewamtouchawareworldaction,
tian2026vtwamvisualtactileworldaction}.
These directions move tactile sensing beyond current-state
conditioning toward both reactive correction and anticipatory
control.

Recent systems increasingly bring prediction and feedback
together. FAWAM and TORL-VLA predict wrench references
for residual or online refinement; OmniVTA compares
predicted and observed tactile features in a high-rate reflex
controller; and UniTacVLA and TouchWorld combine predicted
tactile priors or subgoals with fast action correction
\cite{he2026fawamforceawareworldaction,
zheng2026torlvlatactileguidedonline,
zheng2026omnivtavisuotactileworldmodeling,
zhang2026unitacvlaunifiedtactileunderstanding,
zhou2026touchworld}.
These systems highlight the complementary roles of tactile
foresight and feedback. A common design pattern, however,
is to generate predictive references at the planning or
action-chunk rate and use faster feedback primarily to revise
actions, without re-estimating the predictive reference within
each high-rate tactile update. Building on this direction, our
method learns topology-aware, finger-wise tactile
representations and treats future tactile latents as a
recursively maintained control state, jointly updating them
with the remaining actions as new tactile observations arrive.

% !TEX root = ../AnonymousSubmission2027.tex
\section{Method}

\subsection{Overview}

\begin{figure*}[t]
    \centering
    \includegraphics[width=\textwidth]{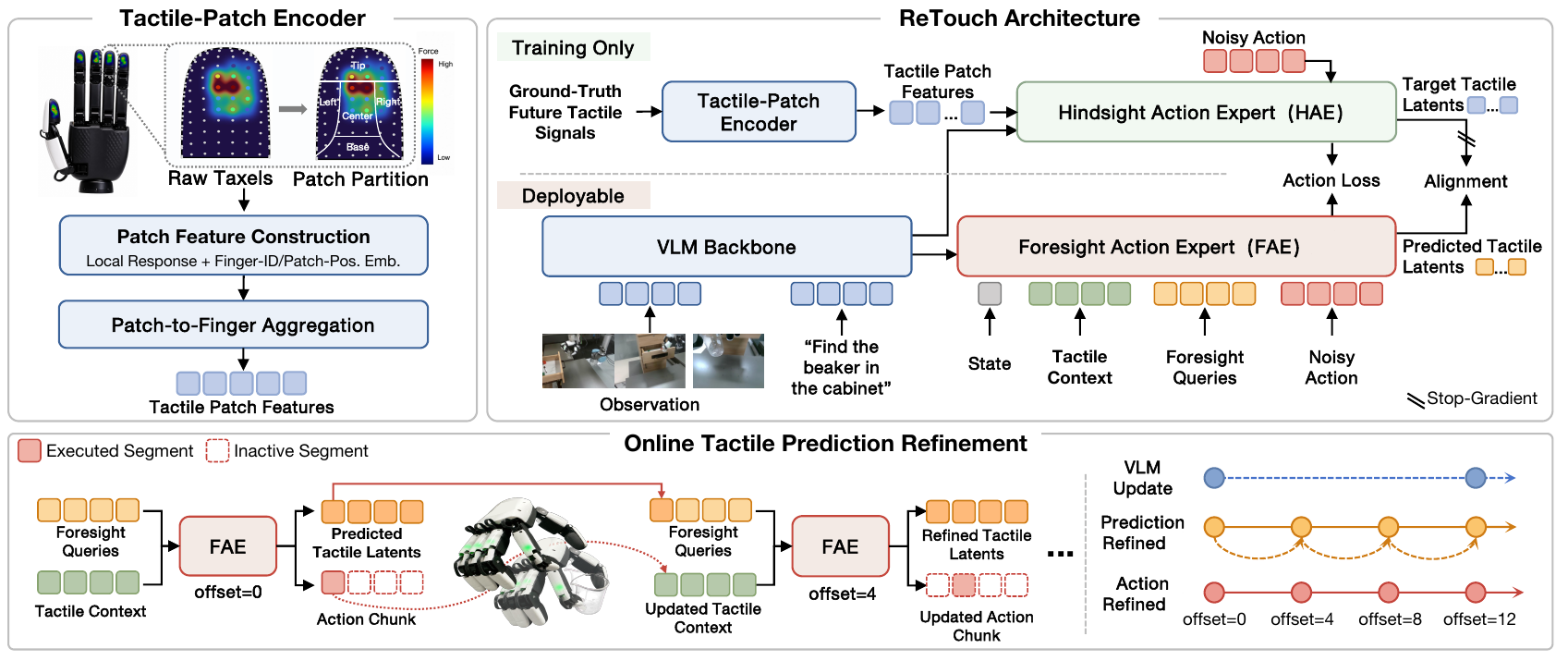}
    \caption{Framework of \systemname{}. The Tactile-Patch Encoder structures
future tactile signals for a training-only Hindsight Action Expert, whose
action-relevant targets supervise the deployable VLM--Foresight Action Expert (FAE)
pathway. At deployment, the FAE refines future tactile latents and the
remaining action chunk using cached VLM context and latest tactile feedback.}
    \label{fig:framework}
\end{figure*}

We propose \systemname, a VLA framework for dexterous manipulation with online-refined tactile prediction. As shown in Fig.~\ref{fig:framework}, \systemname\ decouples low-frequency vision-language reasoning from high-frequency tactile prediction and action updates during execution. At the \(k\)-th low-frequency update, the VLM caches the semantic context \(c_k\), which, together with the chunk-start robot state \(s_k\), conditions subsequent high-frequency control. At each high-frequency call \(t\), the Foresight Action Expert takes the observed tactile context \(\mathcal{T}_t\), comprising the current and historical tactile observations available at time \(t\), along with the future tactile latents carried over from the previous call. It then refines the predicted future tactile latents \(\hat{Z}_t^{+}\) and regenerates the action chunk \(\hat{A}_t\). This procedure is supported by three key designs. First, the Tactile-Patch Encoder organizes five-finger tactile observations into structured tactile patch features, explicitly preserving the finger-wise tactile topology. Second, during training, the Hindsight Action Expert extracts action-relevant future tactile targets from ground-truth future tactile patch features, while the Foresight Action Expert learns to infer the corresponding future tactile latents using only current observations and previous predictions. Finally, during execution, the Foresight Action Expert recursively refines \(\hat{Z}_t^{+}\) from incoming tactile feedback and updates \(\hat{A}_t\), allowing the generated actions to adapt online to changes in contact state.

\subsection{Structured Tactile Patch Encoding}

To preserve finger-wise local contact topology in dexterous-hand tactile signals, the Tactile-Patch Encoder partitions the 120 3D force taxels on each finger into five functional patches corresponding to the tip, center, base, left, and right regions. Let \(\tau_t^{i,j}=(f_x,f_y,f_z)\) denote the 3D force reading of taxel \(j\) on finger \(i\) at time \(t\), and let \(\mathcal{P}_{i,p}\) denote the taxels assigned to patch \(p\). The encoder computes soft contact weights from force magnitudes and aggregates the local tactile responses within each patch:
\begin{equation}
\begin{array}{rcl}
u_t^{i,p}
&=&
\mathrm{Agg}
\left(
\{\tau_t^{i,j}\mid j\in\mathcal{P}_{i,p}\}
\right)
\\[2pt]
&=&
\left[
\bar{\tau}_t^{i,p};
m_t^{i,p};
a_t^{i,p};
q_t^{i,p}
\right].
\end{array}
\end{equation}
Here, \(\bar{\tau}_t^{i,p}\) and \(m_t^{i,p}\) are the contact-gated mean 3D force and the component-wise maximum absolute force, respectively, while \(a_t^{i,p}\) and \(q_t^{i,p}\) denote contact area and contact strength.

The Patch-Informed Tokenizer projects these local responses and adds finger-identity and patch-position embeddings. Each resulting patch feature therefore encodes finger identity, local contact region, and local tactile response. The tokenizer then aggregates the five patch features of each finger into a compact patch-informed finger token. The five resulting tokens, denoted by \(z_t^{\mathrm{tac}}=[z_t^1,\ldots,z_t^5]\), form the structured tactile patch features used by the policy. This design preserves finger-wise local contact topology while exposing only five tactile tokens per frame to the policy.

To encourage each compact finger token to retain recoverable contact information from all five regions, we pretrain the Tactile-Patch Encoder with three prediction heads. From each finger token, these heads predict the relative contact-strength distribution across patches, reconstruct the contact-gated mean 3D force of each patch, and predict its contact state. They are trained with cross-entropy, force regression, and binary cross-entropy, respectively:
\begin{equation}
\mathcal{L}_{\mathrm{TPE}}
=
\lambda_{\mathrm{dist}}\mathcal{L}_{\mathrm{CE}}
+
\lambda_{\mathrm{force}}\mathcal{L}_{\mathrm{force}}
+
\lambda_{\mathrm{contact}}\mathcal{L}_{\mathrm{BCE}}.
\end{equation}
After pretraining, we discard the three prediction heads, initialize the Tactile-Patch Encoder in the policy model with the pretrained weights, and optimize it jointly with the action policy end to end.

\subsection{Learning Action-Relevant Tactile Latents}

Given the structured tactile patch features, we learn future tactile latents that capture contact information relevant to action generation, without reconstructing high-dimensional taxel-level future signals. We use two Action Experts for this purpose: a Hindsight Action Expert (HAE) that constructs supervision targets from ground-truth future tactile signals, and a Foresight Action Expert (FAE) that predicts the corresponding latents from the context available at deployment. The two experts use architecturally identical but parameter-independent Tactile-Patch Encoders. For brevity, we denote their common encoder architecture by \(E_{\phi}\) and omit expert-specific parameter subscripts.

The HAE is a privileged training branch with access to ground-truth future tactile signals. The Tactile-Patch Encoder converts these signals into future tactile patch features, which condition the HAE during action generation. Because the HAE is optimized with the action generation objective, its future tactile representations capture contact information relevant to subsequent actions. At a selected alignment layer \(\ell\), we extract these representations as the hindsight targets:
\begin{equation}
    Z_{t,\ell}^{+,\mathrm{hid}}
    =
    \mathrm{HAE}_{\ell}
    \big(
        c_k,
        s_k,
        \mathcal{T}_t,
        E_{\phi}(\tau_{t:t+H}^{+})
    \big),
\end{equation}
% where \(\tau_{t:t+H}^{+}\) denotes the ground-truth future tactile sequence over prediction horizon \(H\), available only during training.
where \(\tau_{t:t+H}^{+}\) denotes the ground-truth tactile sequence over horizon $H$, available only during training.

The FAE follows the deployable conditioning pathway. It uses learnable Tactile Foresight Queries \(Q^{\mathrm{fore}}\) to infer future tactile latents from the cached semantic context \(c_k\), chunk-start robot state \(s_k\), and observed tactile context \(\mathcal{T}_t\):
\begin{equation}
    \hat{Z}_{t,\ell}^{+}
    =
    \mathrm{FAE}_{\ell}
    \big(
        c_k,
        s_k,
        \mathcal{T}_t,
        Q^{\mathrm{fore}}
    \big).
\end{equation}
We align the predicted latents with the hindsight targets using
\begin{equation}
    \mathcal{L}_{\mathrm{align}}
    =
    d_{\mathrm{cos}}
    \big(
        g_{\psi}(\hat{Z}_{t,\ell}^{+}),
        \mathrm{sg}(Z_{t,\ell}^{+,\mathrm{hid}})
    \big),
\end{equation}
% where \(g_{\psi}\) projects the FAE representations into the HAE latent space, \(d_{\mathrm{cos}}\) denotes the mean cosine distance, and \(\mathrm{sg}(\cdot)\) denotes stop-gradient. This alignment enables the FAE to infer action-relevant future tactile latents from current and historical tactile observations. During deployment, the HAE and the alignment projector are removed, and only the FAE is retained.
where \(g_{\psi}\) projects the FAE representations into the corresponding HAE latent space, \(d_{\mathrm{cos}}\) denotes the mean cosine distance, and \(\mathrm{sg}(\cdot)\) denotes stop-gradient. This alignment enables the FAE to reliably infer action-relevant future tactile latents from current and historical tactile observations. During deployment, both the HAE and the alignment projector are removed, and only the FAE is retained for inference.

\subsection{Online Tactile Prediction Refinement}

Although the Foresight Action Expert can predict future tactile latents, a one-shot prediction may become unreliable over a long action chunk. Slippage, contact shifts, object deformation, and execution errors can cause the initial prediction to diverge from the evolving contact state and potentially misguide subsequent action updates. \systemname\ therefore maintains future tactile latents as execution-time states that are continually refined using incoming tactile feedback. Each refined prediction is then used to revise the remaining actions.

\systemname\ adopts a multi-rate asynchronous execution scheme that separates low-frequency semantic reasoning from high-frequency tactile prediction and action revision. The VLM processes visual observations and language instructions at 9 Hz and caches the resulting semantic context. The Foresight Action Expert runs at 36 Hz and is invoked multiple times between consecutive VLM updates. At each high-frequency call \(t\), it reuses the cached semantic context \(c_k\) and chunk-start robot state \(s_k\), while incorporating the updated tactile context \(\mathcal{T}_t\).

For calls after the initial prediction, we construct the current Tactile Foresight Queries from the latent prefix carried over from the previous call and the learnable queries for the remaining future segments:
\begin{equation}
    \widetilde{Q}_t^{\mathrm{fore}}
    =
    M_t \odot
    \mathrm{sg}\big(\hat{Z}_{t-1}^{+}\big)
    +
    (1-M_t)\odot Q^{\mathrm{fore}},
\end{equation}
where \(M_t\) selects the latent prefix corresponding to the elapsed portion of the action chunk. For the initial call, \(M_t=0\), and the prediction is initialized entirely from \(Q^{\mathrm{fore}}\). The joint prediction and action update is then written as
\begin{equation}
    \big(
        \hat{Z}_{t}^{+},
        \hat{A}_{t}
    \big)
    =
    \mathrm{FAE}_{\theta}
    \big(
        c_k,
        s_k,
        \mathcal{T}_t,
        \widetilde{Q}_t^{\mathrm{fore}}
    \big),
\end{equation}
where \(\hat{Z}_{t}^{+}\) denotes the refined future tactile latents and \(\hat{A}_{t}\) denotes the updated action chunk. After a short action segment is executed, the latest tactile feedback is appended to \(\mathcal{T}_t\) before the next high-frequency call. This allows the Foresight Action Expert to update tactile predictions and actions without rerunning the VLM.

To ensure that each action update uses the tactile prediction produced in the same call, we apply a directional attention mask within the Foresight Action Expert. The Tactile Foresight Queries can attend to the cached semantic context, robot state, observed tactile context, and carried-over latent prefix, but not to action tokens. Action tokens, in contrast, can attend to the updated future tactile latents. The model therefore forms the current tactile prediction before generating the corresponding action chunk, preventing action-token information from leaking into the tactile latents.

Training and deployment follow the same asynchronous refinement scheme. During training, we randomly sample an offset within an action chunk and construct the corresponding tactile context, carried-over latent prefix, and action supervision. From this intermediate execution state, the model learns to update its future tactile latents and the remaining action chunk. During deployment, the Foresight Action Expert is repeatedly invoked at the high-frequency control rate. Each invocation regenerates the action chunk, but only the unexecuted suffix after the current offset is applied. This design closely aligns random-offset training with online tactile prediction and action refinement during execution.

\subsection{Training Objectives}

We train \systemname\ in two stages. We first pretrain the Tactile-Patch Encoder using \(\mathcal{L}_{\mathrm{TPE}}\) defined in Sec.~3.2. After removing the prediction heads, we use the pretrained weights to initialize the parameter-independent encoders in the two Action Experts and optimize them end to end with the policy.

Policy training jointly optimizes action flow matching for both Action Experts, future tactile latent alignment, and the random-offset update objective:
\begin{equation}
    \mathcal{L}_{\mathrm{policy}}
    =
    \mathcal{L}_{\mathrm{act}}^{\mathrm{hid}}
    +
    \mathcal{L}_{\mathrm{act}}^{\mathrm{fore}}
    +
    \lambda_{\mathrm{align}}\mathcal{L}_{\mathrm{align}}
    +
    \mathcal{L}_{\mathrm{update}}.
\end{equation}
% Here, \(\mathcal{L}_{\mathrm{update}}\) applies action and latent-alignment supervision at a randomly sampled intermediate offset. Stop-gradient is applied to the Hindsight targets, preventing the alignment objectives from backpropagating through the Hindsight representations.
Here, \(\mathcal{L}_{\mathrm{update}}\) applies action and latent-alignment supervision at a random intermediate offset. Stop-gradient is applied to Hindsight targets, preventing alignment objectives from backpropagating through the Hindsight representations.
% !TEX root = ../AnonymousSubmission2027.tex
\section{Experiments}

\begin{figure*}[t]
    \centering
    \includegraphics[width=0.95\textwidth]{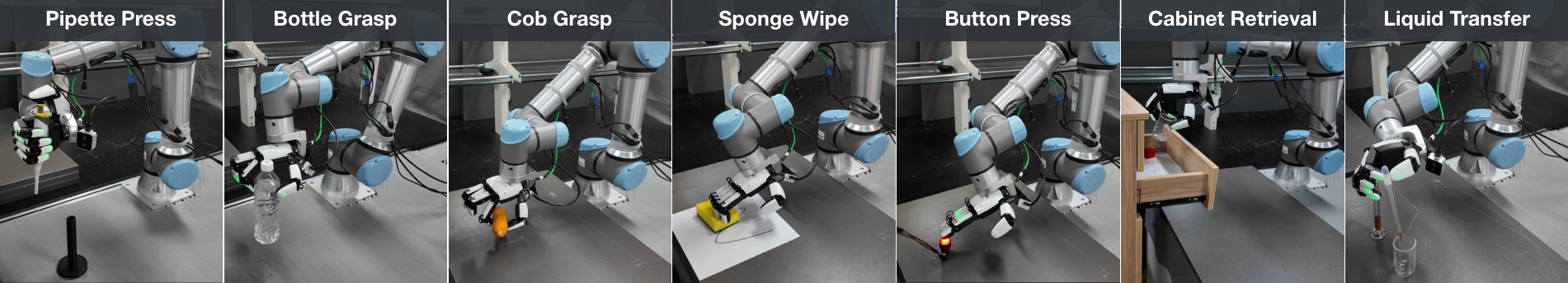}
    \caption{Overview of seven real-world contact-rich dexterous manipulation
tasks on the XHand--UR7e platform, spanning localized pressing, multi-finger
grasping, sustained surface contact, and multi-stage manipulation.}
    \label{fig:experiments}
\end{figure*}

\begin{table*}[t]
\centering

{\small
\setlength{\tabcolsep}{4pt}
\renewcommand{\arraystretch}{1.03}

\begin{tabular*}{\textwidth}
{@{\extracolsep{\fill}}lcccccccc}
\toprule
\raisebox{1.2ex}{\textbf{Method}}
& \shortstack{\textbf{Pipette}\\\textbf{Press}}
& \shortstack{\textbf{Bottle}\\\textbf{Grasp}}
& \shortstack{\textbf{Cob}\\\textbf{Grasp}}
& \shortstack{\textbf{Sponge}\\\textbf{Wipe}}
& \shortstack{\textbf{Button}\\\textbf{Press}}
& \shortstack{\textbf{Cabinet}\\\textbf{Retrieval}}
& \shortstack{\textbf{Liquid}\\\textbf{Transfer}}
& \raisebox{1.2ex}{\textbf{Avg.}} \\
\midrule

RDP
& 46.5 & 65.0 & 45.0 & 50.0 & 5.0 & 10.0 & 0.0 & 31.6 \\

ViTacFormer
& 66.5 & 25.0 & 35.5 & 17.5 & 20.0 & \textbf{67.5} & 0.0 & 33.1 \\

$\pi_{0}$
& 12.5 & 80.0 & 80.0 & 10.0 & 65.0 & 35.5 & 36.5 & 45.6 \\

$\pi_{0}$ + tactile
& 31.5 & 52.5 & 87.5 & 25.0 & 45.0 & 37.5 & 27.5 & 43.8 \\

$\pi_{0.5}$
& 29.5 & 75.0 & 82.5 & 35.0 & 45.0 & 47.5 & 46.0 & 51.5 \\

$\pi_{0.5}$ + tactile
& 47.0 & 60.0 & 92.0 & 20.0 & 60.0 & 42.5 & 50.0 & 53.1 \\

Tactile-VLA
& 84.5 & 82.5 & 95.5 & 62.5 & \textbf{85.0} & 5.0 & 41.5 & 65.2 \\

\midrule

\textbf{\systemname{} (Ours)}
& \textbf{86.0}
& \textbf{95.0}
& \textbf{100.0}
& \textbf{87.5}
& \textbf{85.0}
& 62.5
& \textbf{69.0}
& \textbf{83.6} \\

\bottomrule
\end{tabular*}
}

\caption{Success rates (\%) on seven contact-rich dexterous manipulation
tasks under standard evaluation settings. Each method is evaluated over
20 real-robot rollouts per task using the scoring protocol described above.}
\label{tab:main_results}
\end{table*}

\begin{table}[t]
\centering

{\small
\setlength{\tabcolsep}{4pt}
\renewcommand{\arraystretch}{1.03}

\begin{tabular*}{\columnwidth}
{@{\extracolsep{\fill}}lccccc}
\toprule
\raisebox{1.2ex}{\textbf{Method}}
& \shortstack{\textbf{Height}\\\textbf{Sponge}}
& \shortstack{\textbf{Position}\\\textbf{Cob}}
& \shortstack{\textbf{Lighting}\\\textbf{Liquid}}
& \shortstack{\textbf{Pull}\\\textbf{Bottle}}
& \raisebox{1.2ex}{\textbf{Avg.}} \\
\midrule

ViTacFormer
& 0.0 & 24.5 & 1.5 & 5.0 & 7.8 \\

RDP
& 0.0 & 10.0 & 0.0 & 30.0 & 10.0 \\

Tactile-VLA
& 25.0 & 30.5 & 39.5 & 45.0 & 35.0 \\

$\pi_{0.5}$
& 12.5 & 86.0 & 40.0 & 50.0 & 47.1 \\

$\pi_{0.5}$ + tactile
& 15.0 & 80.5 & 46.5 & 55.0 & 49.3 \\

\midrule

\textbf{\systemname{}}
& \textbf{42.5}
& \textbf{98.5}
& \textbf{66.5}
& \textbf{85.0}
& \textbf{73.1} \\

\bottomrule
\end{tabular*}
}

\caption{Out-of-distribution generalization and disturbance robustness. Evaluated settings include height variation in Sponge Wipe, object-position shift in Cob Grasp, lighting change in Liquid Transfer, and pulling disturbance in Bottle Grasp. Success rates (\%) are reported.}
\label{tab:ood_robustness}
\end{table}

\begin{figure}[t]
    \centering
    \includegraphics[width=\columnwidth]
    {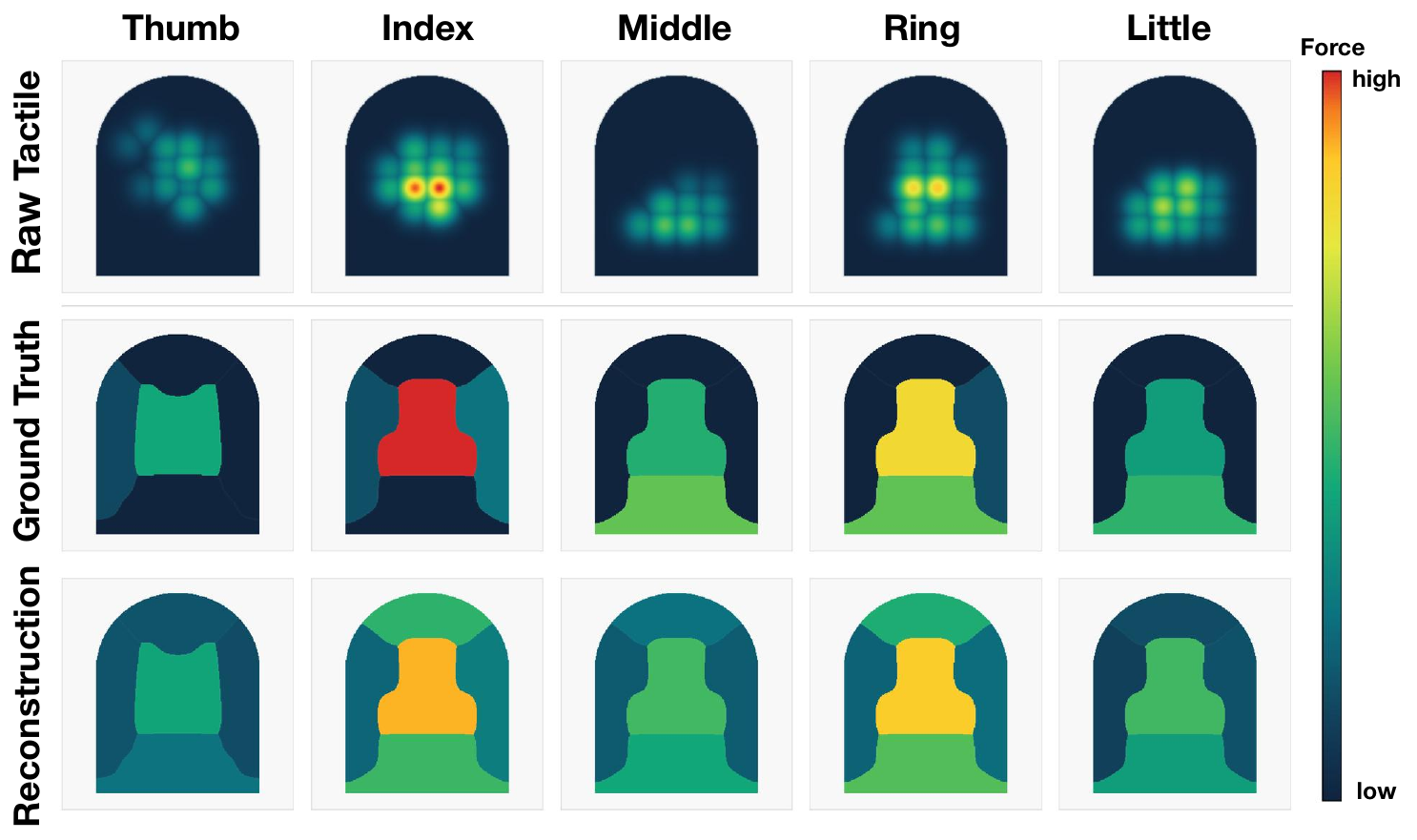}
    \caption{Example of structured tactile patch encoding across the five
fingers. Ground-truth values are patch-wise magnitudes of
contact-strength-weighted mean forces; predicted values are reconstructed
from the corresponding finger-level tactile tokens by the tactile-summary
head.}
    \label{fig:tactile_encoder_analysis}
\end{figure}

\begin{figure}[t]
    \centering
    \includegraphics[width=\columnwidth]{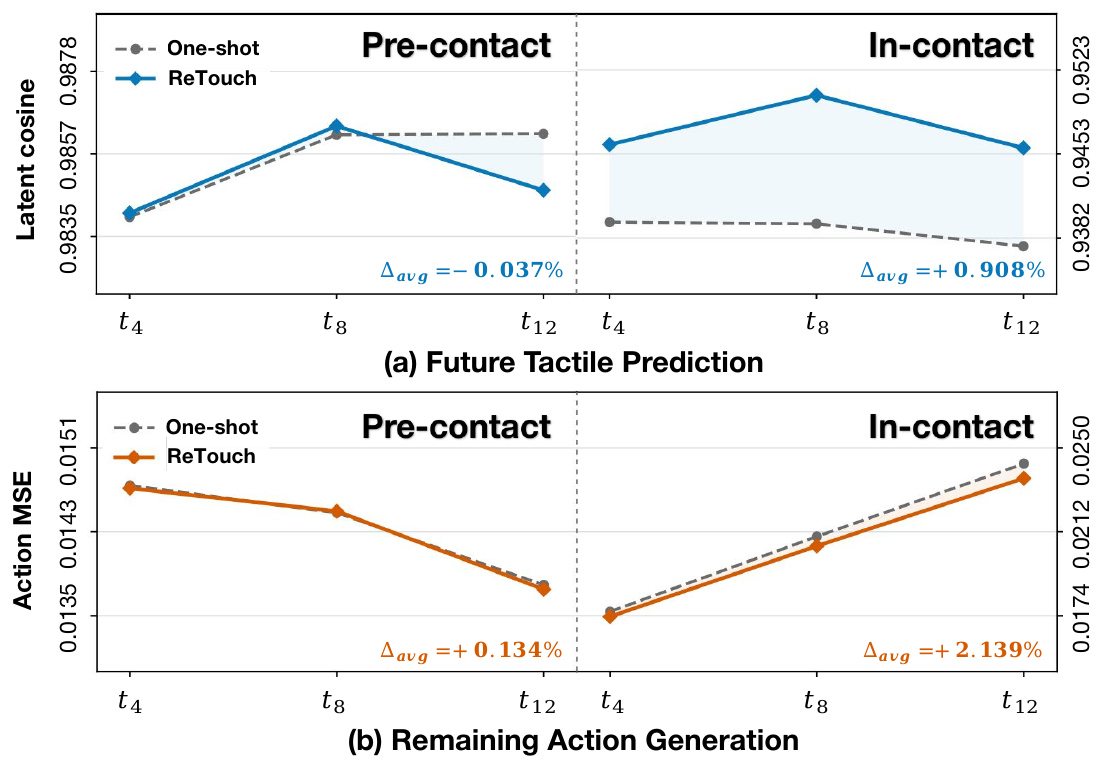}
    \caption{Effect of online refinement of future tactile predictions across
contact phases on future tactile latent similarity (top) and remaining-action
MSE (bottom). Results at $t_0$ are omitted since both methods are identical
before refinement.}
    \label{fig:latent_curve}
\end{figure}

\begin{table*}[t]
\centering

{\small
\setlength{\tabcolsep}{3pt}
\renewcommand{\arraystretch}{1.03}

\begin{tabular*}{\textwidth}
{@{\extracolsep{\fill}}lcccccccc}
\toprule
\raisebox{1.2ex}{\textbf{Variant}}
& \shortstack{\textbf{Pipette}\\\textbf{Press}}
& \shortstack{\textbf{Bottle}\\\textbf{Grasp}}
& \shortstack{\textbf{Cob}\\\textbf{Grasp}}
& \shortstack{\textbf{Sponge}\\\textbf{Wipe}}
& \shortstack{\textbf{Button}\\\textbf{Press}}
& \shortstack{\textbf{Cabinet}\\\textbf{Retrieval}}
& \shortstack{\textbf{Liquid}\\\textbf{Transfer}}
& \raisebox{1.2ex}{\textbf{Avg.} $\uparrow$} \\
\midrule

\textbf{Full \systemname{}}
& 86.0
& \textbf{95.0}
& \textbf{100.0}
& \textbf{87.5}
& 85.0
& \textbf{62.5}
& 69.0
& \makebox[2.2em][r]{\textbf{83.6}}
  \hspace{0em}
  \makebox[3.5em][l]{\textcolor{red!65!black}{(0.0)}} \\

\midrule

w/o intra-chunk refinement
& 74.0
& 62.5
& 95.0
& \textbf{87.5}
& 40.0
& 22.5
& 38.5
& \makebox[2.2em][r]{60.0}
  \hspace{0em}
  \makebox[3.5em][l]{\textcolor{red!65!black}{(-23.6)}} \\

w/o tactile-prediction refinement
& 83.0
& 85.0
& 83.0
& 82.5
& \textbf{90.0}
& 32.5
& 22.5
& \makebox[2.2em][r]{68.4}
  \hspace{0em}
  \makebox[3.5em][l]{\textcolor{red!65!black}{(-15.2)}} \\

w/o future tactile prediction
& 68.0
& 87.5
& 84.5
& 65.0
& 80.0
& 50.0
& 38.0
& \makebox[2.2em][r]{67.6}
  \hspace{0em}
  \makebox[3.5em][l]{\textcolor{red!65!black}{(-16.0)}} \\

non-blocking joint refinement
& \textbf{94.0}
& 80.0
& 98.5
& 85.0
& 65.0
& 45.0
& 63.0
& \makebox[2.2em][r]{75.8}
  \hspace{0em}
  \makebox[3.5em][l]{\textcolor{red!65!black}{(-7.8)}} \\

w/o \Encoder
& 56.5
& 87.5
& 95.5
& 52.5
& 75.0
& 40.0
& \textbf{76.5}
& \makebox[2.2em][r]{69.1}
  \hspace{0em}
  \makebox[3.5em][l]{\textcolor{red!65!black}{(-14.5)}} \\

\bottomrule
\end{tabular*}
}

\caption{Ablation study of \systemname{} across seven contact-rich
dexterous manipulation tasks, reported as success rates (\%). The first
four variants modify online tactile-prediction refinement, while the final
variant removes structured tactile encoding. Parentheses in the Avg. column
show changes from the full model; best results are bold.}
\label{tab:ablation_results}
\end{table*}

\subsection{Experimental Setup}
\noindent{\textbf{Robot Platform.}}
We evaluate \systemname \ on a real XHand--UR7e tactile manipulation platform. The XHand provides dense force-taxel tactile observations from five fingers, with each finger containing 120 three-dimensional taxels. 

\noindent{\textbf{Data and Tasks.}}
We collect 900 tactile-action demonstrations across seven contact-rich
dexterous manipulation tasks. Each demonstration contains visual observations,
language instructions, robot states, robot actions, and dense tactile streams
from the five fingers. The tasks involve diverse physical challenges: (1) Pipette Press, picking up a pipette and precisely pressing its button; (2) Bottle Grasp, grasping transparent water bottles with different weights; (3) Cob Grasp, pinching orange cobs at different heights; (4) Sponge Wipe, wiping marks on a whiteboard with a sponge at different heights; (5) Button Press, pressing small buttons at different heights; (6) Cabinet Retrieval, opening a cabinet drawer to retrieve and place a transparent beaker under visual occlusion; and (7) Liquid Transfer, transferring water from a graduated cylinder to a transparent beaker with a plastic pipette. We train all learning-based methods on the same demonstration set unless otherwise specified.

\noindent{\textbf{Evaluation Protocol and Metrics.}}
We conduct 20 real-robot trials per method on each task. We report normalized task score (\%) as the primary metric.
Single-criterion tasks use binary success, while multi-criterion tasks
assign partial credit through task-specific weighted components.
For brevity, we refer to this metric as success rate throughout. 

\noindent{\textbf{Baselines.}}
We compare ReTouch with seven baselines spanning tactile imitation policies,
pretrained VLA policies, and direct tactile-fusion variants.
ViTacFormer is an ACT-style visuo-tactile imitation policy with
cross-attention tactile fusion and future tactile prediction.
RDP is a slow--fast visual-tactile diffusion policy with
high-frequency tactile-reactive action refinement.
Tactile-VLA is a tactile-aware VLA that incorporates tactile
observations for contact-rich manipulation.
$\pi_0$ and $\pi_{0.5}$ are pretrained VLA
policies fine-tuned on our demonstrations, while
$\pi_0$ + tactile and
$\pi_{0.5}$ + tactile further condition the corresponding
policies on tactile force observations and robot state.

\noindent{\textbf{Implementation Details.}} 
\systemname\ uses a 16-step action horizon and a tactile history window of 9 frames. The vision-language module updates semantic context at 9 Hz, while the Foresight Action Expert runs at 36 Hz. We train a single policy jointly on demonstrations from all seven tasks for 80k steps. The Tactile-Patch Encoder is pretrained for 20k steps with patch-level
structural supervision, frozen for the first 5k policy-training steps, and
then optimized end-to-end. To train asynchronous revision, we randomly sample an intermediate offset from \{4, 8, 12\} within each action chunk. At deployment, only the Foresight Action Expert is retained and invoked at the high-frequency control rate. 

% \subsection{Main Results}
% \input{tabs/main_results}

% \noindent{\textbf{Real-Robot Task Performance}}

% \noindent{\textbf{}}

% main：0725 ，checked

\subsection{Main Results and Analysis}

\noindent{\textbf{Overall Task Performance.}}
Table~\ref{tab:main_results} summarizes the success rates on seven contact-rich dexterous manipulation tasks. \systemname\ achieves the highest macro-average success rate of 83.6\%, outperforming Tactile-VLA, the strongest single baseline on average, by 18.4 percentage points. It ranks first on five tasks and ties for the best result on \textit{Button Press}. The largest gains occur on \textit{Sponge Wipe} and \textit{Liquid Transfer}, where \systemname\ outperforms the strongest task-specific baselines by 25.0 and 19.0 percentage points, respectively. Both tasks involve sustained or repeatedly changing contact, making the observed performance pattern consistent with the design motivation of \systemname. Two observations emerge. First, directly adding tactile observations to pretrained VLA baselines has inconsistent effects: the average success rate of $\pi_0$ decreases from 45.6\% to 43.8\%, whereas that of $\pi_{0.5}$ increases from 51.5\% to 53.1\%. The effects also vary across tasks. Second, existing tactile policies exhibit strong task dependence. Tactile-VLA achieves the strongest baseline macro-average of 65.2\%, whereas ViTacFormer achieves the best result on \textit{Cabinet Retrieval}. This task is the main exception to the overall advantage of \systemname, which achieves 62.5\% compared with 67.5\% for ViTacFormer. Overall, these results show that \systemname\ delivers consistent gains across most contact-rich tasks, with its clearest advantages in settings involving sustained or repeatedly changing contact.

\noindent{\textbf{Structured Tactile Encoding.}}
To assess whether \Encoder\ preserves local contact structure under a compact token budget, we apply the pretrained \Encoder\ to tactile samples from the test set, yielding Tactile Patch Features. We then pass these features through the three prediction heads used during pretraining to recover each patch's contact state, tactile statistics, and contact-strength distribution. % As shown in Fig.~\ref{fig:tactile_encoder_analysis}, the recovered contact regions and strengths for representative samples closely match the ground-truth annotations. 
Fig.~\ref{fig:tactile_encoder_analysis} visualizes the patch-wise force reconstructions, which closely
match the ground-truth values. This agreement indicates that the \Encoder \ compresses dense taxel measurements into a single tactile token per finger while retaining finger identity, local contact location, and contact strength. The resulting representation provides a compact and physically grounded input for subsequent tactile prediction and action generation.

% \noindent{\textbf{tactile prediction修正分析}}

% 为验证推理阶段持续更新触觉预测的作用，我们使用同一个经过异步训练的模型，仅改变其推理策略。\systemname 的 Foresight Action Expert 在每个 action chunk 内利用新观测到的触觉反馈更新未来触觉潜变量，并据此重新预测剩余动作；one-shot 变体在相同时间步重新预测剩余动作，但始终复用 chunk 起始时刻生成的触觉预测。我们使用训练阶段的 hindsight branch 处理真实未来触觉信号，得到目标触觉潜变量（Target Tactile Latents）。在每个更新时刻，我们分别计算预测与目标触觉潜变量后缀的余弦相似度，以及预测与真实动作后缀的物理空间均方误差，以衡量触觉预测和动作生成质量。

% 我们在100条测试轨迹上进行评估，并以被评估手指首次检测到接触的时刻为界，分别统计接触前后的结果。

% 如图~\ref{fig:recursive_revision}所示，接触发生前，触觉信号大多为零，两种推理策略的表现基本一致。接触发生后，\systemname 利用实时触觉反馈持续修正未来预测，使触觉潜变量的平均余弦相似度相对提高0.908\%，并使整体动作误差相对降低2.139\%。动作误差的改善幅度随更新次数增加而扩大。结果表明，固定的初始预测会在接触发生后逐渐偏离实际触觉变化，而基于反馈的持续更新能够使未来触觉预测与实际交互保持一致，并改善后续动作生成。
\noindent{\textbf{Refinement of Tactile Prediction.}} To isolate the effect of continually updating tactile predictions during inference, we use the same model trained with asynchronous revision and vary only its inference strategy. Within each action chunk, the Foresight Action Expert of \systemname{} incorporates newly observed tactile feedback to update its future tactile latents and re-predict the remaining actions. The one-shot variant re-predicts the remaining actions at the same timesteps but reuses the tactile prediction generated at the beginning of the chunk. We obtain the Target Tactile Latents by processing the ground-truth future tactile signals with the hindsight branch used during training. At each update, we evaluate tactile prediction using the cosine similarity between the suffixes of the predicted and target latent sequences, and action generation using the physical-space MSE between the predicted and ground-truth action suffixes. We evaluate both strategies on 100 test trajectories. Each trajectory is divided into pre-contact and in-contact phases at the first timestep when the finger registers contact. As shown in Fig.~\ref{fig:latent_curve}, tactile signals are mostly zero before contact, and the two inference strategies perform similarly. 
After contact, \systemname{} continually revises its future predictions using tactile feedback, improving the mean latent cosine similarity by 0.908\% and reducing the overall action error by 2.139\% relative to the one-shot variant. The reduction in action error grows with successive updates. These results suggest that a fixed initial prediction becomes increasingly misaligned with the evolving tactile state after contact. In contrast, feedback-driven updates keep future tactile predictions aligned with the ongoing interaction and thereby improve subsequent action generation.

\subsection{Robustness to Perturbations}

% 我们进一步在四类chellenging settings 下 评估 \systemname{}，包括操作高度变化、物体位置偏移、光照变化和抓取后的外力拉拽。如表~\ref{tab:ood_robustness} 所示，\systemname{} 平均成功率达到 \(73.1\%\)，相比平均性能最强的基线 \(\pi_{0.5}\)+tactile 提高了 23.8 个百分点。在改变接触几何关系的高度变化和物体位置偏移条件下，\systemname{} 均获得最高成功率。在主要改变视觉观测的光照变化条件下，\systemname{} 相比最强基线的提高 20.0 个百分点，表明触觉反馈能够在视觉观测发生变化时提供互补的局部接触信息。最大的性能提升出现在抓取后的外力拉拽条件下，\systemname{}相比最强基线的 \(55.0\%\) 提高 30.0 个百分点。该扰动直接改变抓取建立后的手指受力分布和物体稳定性，因此这一结果与执行过程中持续利用触觉反馈修正未来预测的设计动机一致。总体而言，这些结果表明，\systemname{} 能够在视觉变化、接触几何变化和执行中的物理扰动下保持较强的操作性能
We further evaluate \systemname{} under four challenging settings: variations in manipulation height, object-position shifts, lighting changes, and external pulling disturbances after grasp. As shown in Table~\ref{tab:ood_robustness}, \systemname{} achieves an average success rate of 73.1\%, outperforming \(\pi_{0.5}\)+tactile, the strongest baseline on average, by 23.8 percentage points. Under height variations and object-position shifts, both of which alter the contact geometry, \systemname{} achieves the highest success rate. Under lighting changes, which primarily affect visual observations, \systemname{} outperforms the strongest baseline by 20.0 percentage points, suggesting that tactile feedback provides complementary local contact information when visual observations change. The largest gain occurs under post-grasp pulling disturbances, where \systemname{} achieves 85.0\%, exceeding the strongest baseline at 55.0\% by 30.0 percentage points. This disturbance alters the finger-force distribution and object stability after grasp establishment, a result consistent with the design motivation of continually refining future tactile predictions using execution-time feedback. Overall, these results show that \systemname{} maintains strong performance under visual changes, contact-geometry variations, and physical disturbances during execution.

\subsection{Ablation Studies}

% 表~\ref{tab:ablation_results} 考察了在线触觉预测修正和结构化触觉编码的贡献。去除触觉预测和动作的动作块内修正会造成最大性能下降，使平均成功率降低23.6个百分点，表明仅在动作块开始时生成预测和动作难以适应不断变化的接触状态。保留动作更新但固定初始触觉预测时，平均成功率为 \(68.4\%\)，相比完整模型低15.2个百分点；完全移除未来触觉预测则得到 \(67.6\%\)。二者仅相差0.8个百分点，说明固定的一次性触觉预测收益有限，主要提升来自利用最新反馈持续修正未来预测。非阻塞变体指的是在不中断当前动作执行的情况下并行更新触觉预测和动作块，因此修正结果会延迟约一个动作步后生效。该变体的平均成功率为 \(75.8\%\)，相比完整模型低7.8个百分点，表明更新延迟也会削弱在线修正的效果。移除 Tactile-Patch Encoder 后，平均成功率下降14.5个百分点至 \(69.1\%\)，说明保留手指身份和局部接触区域对整体性能具有重要作用。总体而言，未来触觉预测需要在执行过程中持续更新才能有效支持动作修正，而结构化触觉编码提供了互补的性能收益。
Table~\ref{tab:ablation_results} examines the contributions of online tactile-prediction refinement and structured tactile encoding. Removing intra-chunk refinement of both tactile predictions and actions causes the largest degradation, reducing the average success rate by 23.6 percentage points. This result indicates that predictions and actions generated only at the beginning of an action chunk cannot reliably adapt to evolving contact states. Retaining action updates while keeping the initial tactile prediction fixed yields 68.4\%, 15.2 percentage points below the full model, whereas removing future tactile prediction entirely yields 67.6\%. The 0.8-point difference suggests that a fixed one-shot tactile prediction provides limited benefit and that the main gain comes from continually refining future tactile predictions with the latest feedback. The non-blocking variant updates tactile predictions and action chunks in parallel without pausing execution, causing each refinement to take effect approximately one action step later. It achieves 75.8\%, 7.8 percentage points below the full model, suggesting that delayed updates weaken online refinement. In the w/o Tactile-Patch Encoder variant, we flatten the raw tactile signals and encode them with an MLP before fusion, removing the explicit finger-wise and local-region structure while keeping the rest of the model unchanged. This variant reduces the average success rate by 14.5 percentage points to 69.1\%, supporting the benefit of preserving finger identity and local contact regions in tactile representations.
% !TEX root = ../AnonymousSubmission2027.tex
\section{Conclusion}

% 我们提出了 ReTouch，一种面向灵巧操作的有Online-Refined Tactile Prediction的视觉—语言—动作模型。该模型保留逐指局部接触拓扑结构，并在执行过程中持续在线修正未来触觉预测和动作块。在 XHT-Bench 的七项真实世界任务中，ReTouch 在标准测试和挑战性测试条件下的平均成功率分别比最强基线提高了18.4和23.8个百分点。通过实验验证表明，结构化触觉编码与及时的块内修正均是性能提升的关键。相比固定的一次性预测，反馈驱动的持续更新能够更好地适应接触变化并改善动作生成。因此，触觉预测应作为随执行反馈持续更新的控制状态，而非一次性预测。

We introduce ReTouch, a vision-language-action model for dexterous manipulation that preserves finger-wise local contact topology and jointly refines future tactile predictions and action chunks during execution. Across seven real-world XHT-Dataset tasks, ReTouch improves average success rate over the strongest baseline by 18.4 and 23.8 percentage points under standard and challenging settings, respectively. Experiments identify structured tactile encoding and timely intra-chunk refinement as key contributors. Compared with fixed one-shot predictions, feedback-driven updates better accommodate contact changes and improve action generation. Tactile prediction is therefore most effective as a control state updated by execution feedback rather than a fixed forecast.


% \clearpage
\bibliography{aaai2027}

\clearpage
\appendix

% =========================================================
\section{Implementation Details}
\label{app:implementation}
% =========================================================

\subsection{Model Architecture}
\label{app:model_io}

\paragraph{Policy interface.}
ReTouch is initialized from a pretrained $\pi_0$ checkpoint and fine-tuned
during policy training. Its visual input consists of one wrist-mounted view
and two fixed external views. All views are resized to $224\times224$ using
aspect-ratio-preserving padding and normalized to $[-1,1]$. The policy also
receives a task instruction and an 18-dimensional proprioceptive state
comprising six UR7e joint positions and twelve XHand joint positions, and
outputs absolute joint-position commands in the same 18-dimensional space. Each tactile frame contains 120 three-axis taxels per finger,
$\tau_t\in\mathbb{R}^{5\times120\times3}$. The tactile history contains the
preceding nine frames; together with the current frame, this forms the
ten-frame input window
\begin{equation}
\mathcal{T}_t =
[\tau_{t-9},\ldots,\tau_t]
\in\mathbb{R}^{10\times5\times120\times3}.
\label{eq:tactile_window}
\end{equation}
At episode boundaries, unavailable history is filled by repeating the
earliest available frame. At refinement offset $o$, the window is updated to
$\mathcal{T}_{t+o}=[\tau_{t+o-9},\ldots,\tau_{t+o}]$.

\paragraph{Action refinement.}
The policy predicts a 16-step action chunk
$\hat{A}^{(o)}_t\in\mathbb{R}^{16\times18}$. The chunk regenerated at offset
$o\in\{4,8,12\}$ remains aligned with the start of the original chunk; its
executed prefix is discarded, and only
$\hat{A}^{(o)}_t[o{:}16]$ is applied.

\subsection{Tactile-Patch Encoder}
\label{app:tactile_patch_encoder}

\begin{figure}[t]
    \centering
    \includegraphics[width=0.7\columnwidth]{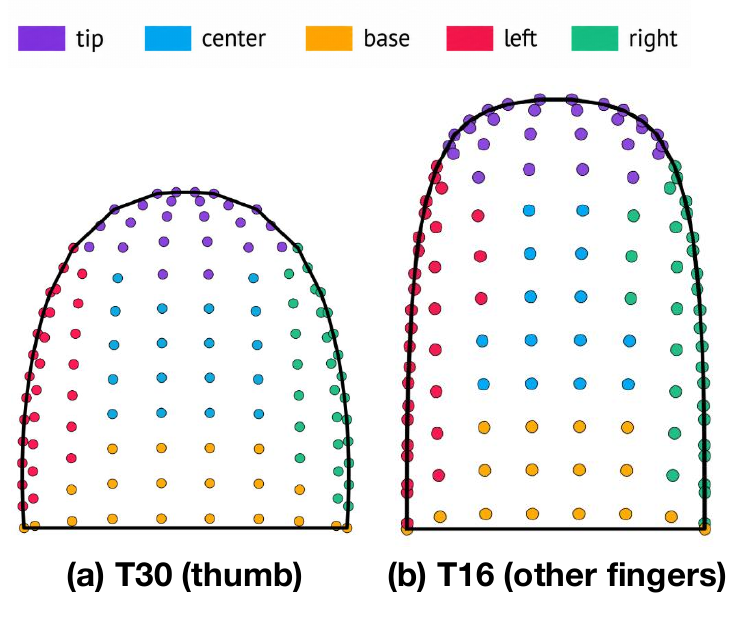}
    \caption{Fixed XHand taxel-to-patch maps for
    (a) the T30 thumb sensor and
    (b) the T16 sensors used by the other four fingers.
    Each color denotes one spatial patch.}
    \label{fig:xhand_patch_maps}
\end{figure}

\paragraph{Patch assignment and representation.}
The thumb uses a fixed T30 taxel-to-patch map, while the other four fingers
share a fixed T16 map, as visualized in Fig.~\ref{fig:xhand_patch_maps}.

Taxel readings are normalized independently along each force axis using
training-set statistics. For patch $p$ with taxel index set $\mathcal I_p$,
the soft contact weights are
\begin{equation}
w_j=\sigma\!\left(
\frac{\|\mathbf f_j\|_2-\theta_c}{\gamma_c}
\right),
\qquad
W_p=\sum_{j\in\mathcal I_p}w_j .
\label{eq:tpe_gate}
\end{equation}
The patch statistics used in the main paper are computed as
\begin{equation}
\begin{aligned}
\bar{\mathbf f}_p &=
\frac{\sum_{j\in\mathcal I_p}w_j\mathbf f_j}
{\max(W_p,\epsilon)},&
\mathbf m_p &=
\operatorname*{max}_{j\in\mathcal I_p}|\mathbf f_j|,\\
a_p &= \frac{W_p}{|\mathcal I_p|},&
q_p &= \max_{j\in\mathcal I_p}\|\mathbf f_j\|_2 ,
\end{aligned}
\label{eq:tpe_statistics}
\end{equation}
where $\mathbf m_p$ is taken componentwise. We use $\theta_c=0.5$ during
encoder pretraining and $\theta_c=1.0$ during policy training, with
$\gamma_c=0.5$ and $\epsilon=10^{-6}$.
\paragraph{Spatial and temporal aggregation.}
The descriptor projection is an
$8\!\rightarrow\!256\!\rightarrow\!1024$ SiLU MLP. Let $\mathbf h_p$ denote
an embedded patch feature. The exact patch pooling is
\begin{equation}
\begin{aligned}
e_p &=
g\!\left(\operatorname{SiLU}(\mathbf h_p)\right)
+\log(a_p+\epsilon),\\
\alpha_p &=
\frac{\exp(e_p)}
{\sum_{r=1}^{5}\exp(e_r)},
\qquad p=1,\ldots,5,\\
\mathbf z &=
\operatorname{LN}\!\left(
\sum_{p=1}^{5}\alpha_p\mathbf h_p
\right).
\end{aligned}
\label{eq:tpe_pooling}
\end{equation}
where $g$ is a learned scalar projection and $\alpha_p$ is the normalized
weight of patch $p$. Learned scalar weights pool the
first nine tactile frames per finger, while the current-frame tokens are
retained with separate history/current type embeddings, producing ten input
tokens. For hindsight encoding, four consecutive four-step segments produce
20 future-tactile tokens.

\paragraph{Encoder pretraining.}
For the pretraining objective defined in the main paper, the contact target is
$y_p=\mathbf{1}(q_p>\theta_c)$, and the distribution target normalizes
$y_pq_p$ across patches, using a uniform distribution when no patch is active.
We set the contact-loss weight to $0.5$ and use the balanced force term
\begin{equation}
\mathcal L_{\mathrm{force}}=
\mathcal L_{\mathrm{active}}
+0.5\,\mathcal L_{\mathrm{magnitude}}
+0.25\,\mathcal L_{\mathrm{inactive}},
\label{eq:tpe_balanced_force}
\end{equation}
where the three Smooth-$L_1$ terms regress active-patch forces,
active-patch force magnitudes, and zero force on inactive patches,
respectively; active and inactive patches are normalized separately.
Two history and two future frames are randomly sampled per training example.

\subsection{Hindsight--Foresight Learning and Online Refinement}
\label{app:online_refinement_details}

\paragraph{Latent alignment.}
The HAE and FAE use separate 18-layer Transformers with hidden width 1024
and eight attention heads. At alignment layer $\ell=12$, the HAE states at
the 20 future-tactile positions defined in
Sec.~\ref{app:tactile_patch_encoder} serve as stop-gradient targets. The
corresponding FAE states are mapped by a
$1024\!\rightarrow\!1024\!\rightarrow\!1024$ SiLU projector and aligned
using mean cosine distance with weight $0.1$.

\paragraph{Refinement scheduling.}
Full ReTouch uses blocking refinement: at each update offset, execution holds
the last command until the updated action chunk is returned. The non-blocking
ablation uses the same checkpoint, continues executing the current chunk
during deployment, and replaces only the suffix remaining when the result
arrives; expired results are discarded.

\subsection{Training and Inference Details}
\label{app:training_inference}

\paragraph{Optimization and augmentation.}
Encoder pretraining and policy training use global batch sizes 32 and 8,
respectively. We use AdamW with gradient clipping at 1.0, 1k warmup updates,
and cosine learning-rate decay from $2.5\times10^{-5}$ to
$2.5\times10^{-6}$. Training runs on eight NVIDIA A800 GPUs using bfloat16
mixed precision. All three views receive color jitter with brightness,
contrast, and saturation strengths of $0.3$, $0.4$, and $0.5$, respectively.
One designated external view, referred to as the base view, additionally
receives a random crop retaining 95\% of the image area and a random rotation within $\pm5^\circ$.

\paragraph{Action flow matching.}
For a normalized action chunk $a$, we sample
$\varepsilon\sim\mathcal{N}(0,I)$ and
$t=0.001+0.999\tilde t$, where
$\tilde t\sim\mathrm{Beta}(1.5,1)$, and define
$x_t=t\varepsilon+(1-t)a$ and $u=\varepsilon-a$.
The HAE and FAE receive the same $(x_t,t)$ and regress $u$ using mean-squared
error. Inference uses ten explicit-Euler steps from $t=1$ to $0$, with all
refinement calls in the same chunk reusing its initial action-noise sample.

\paragraph{Runtime latency.}
Table~\ref{tab:latency_results} reports mean model-only latency on an NVIDIA
RTX~5090, excluding communication, command execution, and controller-side
scheduling. With one full VLM--FAE pass and three cached-context FAE
refinements per cycle, the measured latencies correspond to 9.02 complete
cycles and 36.08 FAE passes per second, reported as 9~Hz and 36~Hz in the
main text.

% \begin{table}[t]
% \centering
% \caption{Inference latency of \systemname \ during deployment. Latency is reported per control cycle or asynchronous update stage.}
% \label{tab:latency_results}
% \footnotesize
% \setlength{\tabcolsep}{6pt}
% \begin{tabular}{lc}
% \toprule
% \textbf{Component}
% & \textbf{Latency (ms)} \\
% \midrule
% VLM cache construction
% & -- \\
% Patch-aware tactile encoder
% & -- \\
% Action expert update
% & -- \\
% Async correction step
% & -- \\
% \bottomrule
% \end{tabular}
% \end{table}

% \begin{table}[t]
% \centering

% \label{tab:latency_results}
% \footnotesize
% \setlength{\tabcolsep}{6pt}
% \renewcommand{\arraystretch}{1.08}

% \begin{tabular}{@{}lc@{}}
% \toprule
% \textbf{Inference stage}
% & \textbf{Latency (ms)} \\
% \midrule
% Initial VLM--FAE pass
% & 52.25 \\
% Cached-context FAE refinement
% & 19.54 \\
% \bottomrule
% \caption{Deployment inference latency of \systemname{} on an NVIDIA RTX~5090.
% Latency denotes average model-compute time and excludes communication, command
% execution, and controller-side scheduling overhead.}
% \end{tabular}
% \end{table}

\begin{table}[t]
\centering

\footnotesize
\setlength{\tabcolsep}{6pt}
\renewcommand{\arraystretch}{1.08}

\begin{tabular}{@{}lc@{}}
\toprule
\textbf{Inference stage}
& \textbf{Latency (ms)} \\
\midrule
Initial VLM--FAE pass
& 52.25 \\
Cached-context FAE refinement
& 19.54 \\
\bottomrule
\end{tabular}

\caption{Deployment inference latency of \systemname{} on an NVIDIA RTX~5090.
Latency denotes average model-compute time and excludes communication, command
execution, and controller-side scheduling overhead.}
\label{tab:latency_results}

\end{table}
% =========================================================
\section{Real-World Experimental Setup}
\label{app:experimental_setup}
% =========================================================

\subsection{Robot Platform and Data Collection}
\label{app:robot_data_collection}

\begin{figure}[!t]
    \centering
    \includegraphics[width=0.98\columnwidth]{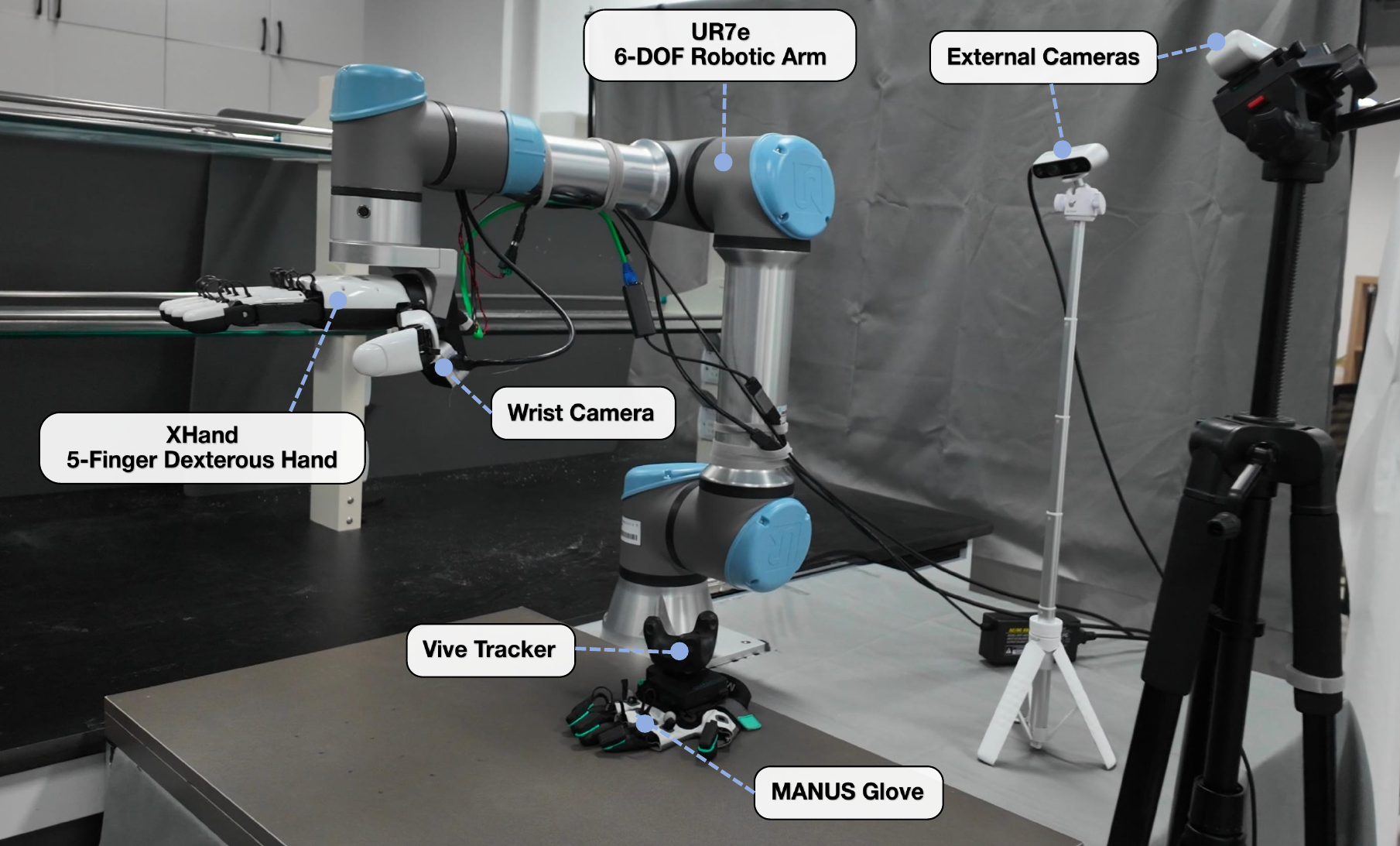}
    \caption{Real-world system used for demonstration collection and policy
    evaluation. Wrist motion is captured with a VIVE tracker, finger motion
    with a MANUS glove, and visual observations with one wrist-mounted and two
    fixed external RGB cameras.}
    \label{fig:real_world_setup}
\end{figure}

As shown in Fig.~\ref{fig:real_world_setup}, the platform consists of a UR7e arm, an XHand five-finger dexterous hand, one
wrist-mounted RGB camera, and two fixed external RGB cameras. During
teleoperation, a VIVE tracker provides the wrist-pose target, while a MANUS
glove records finger motion.

Before each run, the robot and task objects are returned to their prescribed
initial configurations, and the tactile sensors are zeroed. We retain
successful demonstrations and discard demonstrations with interrupted
execution or corrupted recordings. Except for the prescribed pulling
disturbance in Sec.~\ref{app:challenging_settings}, no corrective human
intervention is used during policy rollouts.

\subsection{Dataset Composition and Partitions}
\label{app:dataset_composition}

XHT-Dataset contains 900 successful demonstrations across the seven tasks.
We hold out 100 trajectories as a held-out offline test set, excluding them
from both Tactile-Patch Encoder pretraining and ReTouch policy training. The
remaining 800 trajectories form the common training pool. The held-out offline
test set is used only for the supplementary diagnostics in
Sections~\ref{app:tpe_pretraining_ablation} and
\ref{app:recursive_refinement_analysis}. The split is defined at the trajectory level before tactile windows
and action chunks are sampled. Both diagnostic analyses use the same held-out offline test set.

ReTouch, Tactile-VLA, and the $\pi$-family baselines are trained jointly across
all seven tasks. RDP and ViTacFormer retain their task-specific training
formulations and draw from the corresponding task-specific subsets of the
common training pool. ViTacFormer further constructs its own method-specific
validation split within each task's training data for checkpoint selection and
does not use the 100 held-out test trajectories.

\subsection{Evaluation Tasks and Scoring Criteria}
\label{app:task_scoring}

\begin{figure*}[!t]
    \centering
    \includegraphics[width=\textwidth]{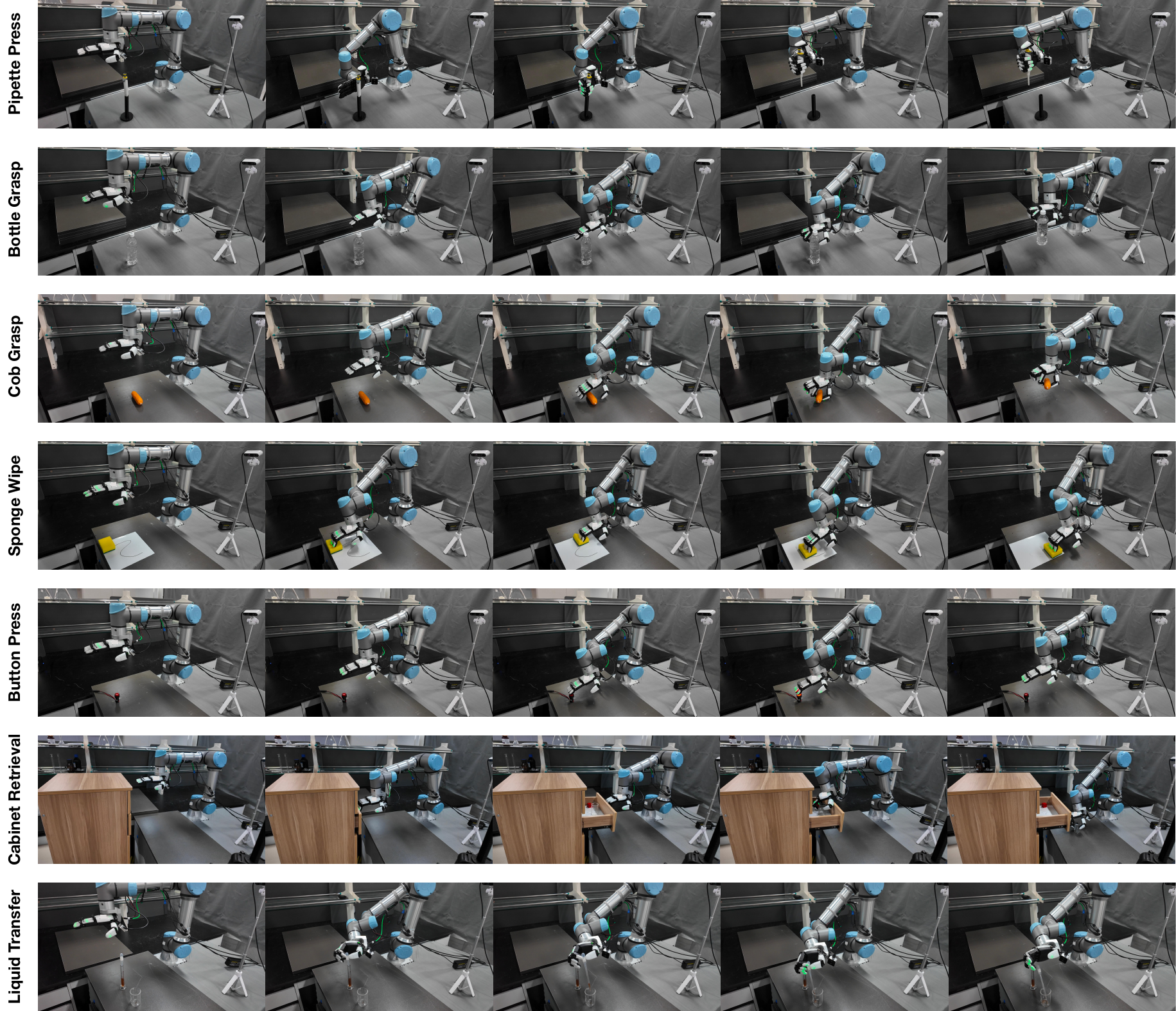}
    \caption{Representative execution sequences for the seven real-world
    tasks. Each row shows five temporal snapshots from initialization to task
    completion.}
    \label{fig:task_progressions}
\end{figure*}

Each method is evaluated on 20 real-robot rollouts per task using the same
predefined initial-configuration distribution. The first five tasks use a
60-s timeout, whereas Liquid Transfer and Cabinet Retrieval use a 120-s
timeout. Within the timeout, a policy may retry or regrasp without human
assistance. Protective stops and unrecoverable out-of-workspace failures
terminate the rollout, and the score reflects the stages completed before
termination.

Each rollout receives a normalized task score in $[0,1]$. Button Press uses a
binary criterion: the rollout receives a score of 1.0 when the button indicator
light turns on. The remaining tasks use additive stage-based scores. Figure~\ref{fig:task_progressions} shows representative execution sequences
for the seven tasks.

Pipette Press assigns 0.4 for grasping the pipette, an additional 0.3 when the
button is depressed through at least one-third but less than one-half of its
full travel, and another 0.3 when the depression reaches at least one-half of
the full travel. Bottle Grasp assigns 0.5 for lifting the bottle at least
15~cm above its initial support surface and 0.5 for maintaining the lifted
grasp for 3~s. Cob Grasp assigns 0.4 for lifting the cob, 0.3 for attaining a
stable grasp posture, and 0.3 for holding it without slip for 3~s.

Sponge Wipe assigns 0.5 when part of the target marks is removed but visible
black traces remain, and an additional 0.5 when no visible black trace remains.
Liquid Transfer assigns 0.3 for grasping the dropper, 0.3 for aspirating
liquid, and 0.4 for positioning the dropper above the beaker and dispensing
the aspirated liquid into it. Cabinet Retrieval assigns 0.5 for opening the
middle drawer and 0.5 for retrieving and placing the beaker.

A single evaluator applies the fixed rubric on site. The task score is averaged
over 20 rollouts, and the overall score is the macro-average across the seven
tasks. Accordingly, ``Success Rate'' in the main paper denotes the mean
normalized task score (\%); except for Button Press, it is a graded
task-completion score rather than binary completion frequency.

\subsection{Challenging Evaluation Settings}
\label{app:challenging_settings}

\begin{figure*}[!t]
    \centering
    \includegraphics[width=\textwidth]{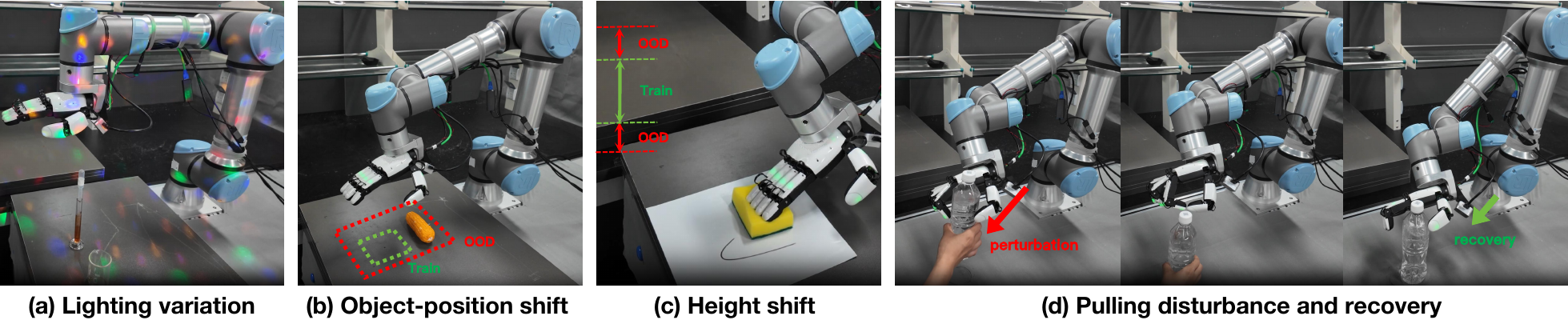}
    \caption{Challenging evaluation settings:
(a) Liquid Transfer under dynamic projector-generated illumination,
(b) Cob Grasp at held-out object positions,
(c) Sponge Wipe at unseen manipulation heights, and
(d) Bottle Grasp with a downward pulling disturbance applied after stable
lifting, followed by autonomous grasp recovery.}
    \label{fig:challenging_settings}
\end{figure*}

The four challenging settings are illustrated in
Fig.~\ref{fig:challenging_settings}:
Sponge Wipe at held-out heights of 2.5 and 5~cm, compared with training heights
of 7.5, 9, and 10.5~cm; Cob Grasp at held-out positions outside the training
region; Liquid Transfer under several dynamic illumination patterns generated
by a projector; and Bottle Grasp under a pulling disturbance. For the last
setting, after the bottle is stably lifted, the same experimenter pulls it
downward until it is dislodged from the hand. The rollout then continues, and
the policy is allowed to recover the grasp autonomously. Each reported method
is evaluated on 20 rollouts in each challenging setting using the scoring
criteria defined in Sec.~\ref{app:task_scoring}.

Among the $\pi$-family baselines, the main-paper robustness evaluation reports
$\pi_{0.5}$ and $\pi_{0.5}+\mathrm{tactile}$, which outperform their
corresponding $\pi_0$ variants on average under the standard setting. All
tactile-specific baselines and ReTouch are retained.

% =========================================================
\section{Implementation Details of Baselines}
\label{app:baseline_details}
% =========================================================

All baselines are evaluated on the XHand--UR7e platform using the common
18-dimensional absolute joint-position command space and a 16-step action
chunk. We adapt each method to the XHand sensor
layout while preserving its core architecture and execution scheme. The exact
visual, proprioceptive, and tactile inputs, training procedures, and
checkpoint-selection rules are specified below.

\paragraph{ViTacFormer.}
Our ViTacFormer reproduction follows the original ACT-based implementation and
is trained separately for each task. The model receives 16-frame histories of
the proprioceptive state and tactile observations. At each frame, the tactile
observation comprises the stored 3D resultant-force vector from each of the
five fingers together with its corresponding temporal difference, yielding a
30-dimensional representation. The future-touch branch predicts a 16-step
sequence of future tactile representations in the same 30-dimensional form,
and the resulting future-touch tokens condition the action Transformer during
inference. The ACT decoder produces a 16-step chunk of absolute joint-position
commands.

Each task-specific policy is trained for 30k updates with batch size 64. We use
AdamW with learning rates of $10^{-4}$ for the policy and $10^{-5}$ for the
visual backbone, and select the checkpoint with the lowest validation loss.
During deployment, the policy uses ACT temporal ensembling at each control
step.

\paragraph{Reactive Diffusion Policy.}
For RDP, we retain its task-specific two-stage AT--LDP training and slow--fast
execution scheme. Each finger's 360-dimensional raw tactile vector is reduced
to 20 dimensions using PCA, yielding a 100-dimensional tactile representation
across the five fingers. The PCA basis is fitted only on the pooled training
data from all seven tasks and shared across all task-specific policies.

For each task, the Asymmetric Tokenizer (AT) is trained for 100 epochs with
batch size 64 and learning rate $10^{-3}$. The Latent Diffusion Policy (LDP)
is then trained for 200 epochs with batch size 32 and learning rate $10^{-4}$
using the final AT checkpoint. Deployment uses the EMA weights of the LDP.
At each control step, the tactile-conditioned AT decodes the next action
command using the latest tactile observation, whereas the LDP refreshes the
slow latent once every 13 control steps. No language conditioning is used.

\paragraph{Tactile-VLA.}
The policy is initialized from the pretrained $\pi_0$ checkpoint. We adapt its
tactile input to a ten-frame XHand taxel history of shape $[10,5,120,3]$,
corresponding to time, fingers, taxels, and force axes, respectively. The
flattened history is mapped by an MLP to a single 2048-dimensional tactile
prefix token that conditions the VLA policy. In parallel, 16 auxiliary
force-query tokens are used to predict future tactile information over the
16-step action horizon.

We train a single policy jointly across all seven tasks for 80k updates with
batch size 8, bfloat16 computation, and the same AdamW warmup--cosine
learning-rate schedule as ReTouch. Evaluation uses the final checkpoint and
ten-step Euler integration. The auxiliary tactile-prediction MSE has weight
0.2 and contributes only to the training objective. Its predictions are not
used to condition action generation or during deployment.

\paragraph{$\pi_0$ and $\pi_0$+tactile.}
Both variants are initialized from the pretrained $\pi_0$ checkpoint. The
standard $\pi_0$ variant receives the three RGB views, the language instruction,
and the continuous 18-dimensional proprioceptive state. In
$\pi_0+\mathrm{tactile}$, the current $5\times120\times3$ tactile frame is
flattened and concatenated with the proprioceptive state.

\paragraph{$\pi_{0.5}$ and $\pi_{0.5}$+tactile.}
Both variants are initialized from the pretrained $\pi_{0.5}$ checkpoint. In
the standard $\pi_{0.5}$ pathway, the proprioceptive state is discretized and
serialized into the textual state field. For
$\pi_{0.5}+\mathrm{tactile}$, the current flattened tactile frame is
concatenated with the proprioceptive state before discretization. The normalized scalars are each assigned to one of 256 bins, serialized into
the textual state field, and processed by the language-tokenization pathway.

All four
$\pi$ baselines are jointly fine-tuned across all seven tasks for 80k updates
with batch size 8, bfloat16 computation, and the same AdamW warmup--cosine
learning-rate schedule as ReTouch. Evaluation uses the final checkpoints and
ten-step Euler integration.

% =========================================================
\section{Additional Experiments and Analyses}
\label{app:additional_experiments}
% =========================================================

\subsection{Real-Robot Ablation Variants}
\label{app:real_robot_ablations}
\noindent{\textbf{Full ReTouch.}}
The full model uses structured tactile tokens, future tactile prediction,
recursive refinement of predicted tactile latents, and blocking updates at
offsets 4, 8, and 12.

\noindent{\textbf{The w/o intra-chunk refinement variant.}}
The policy executes the full action chunk generated at the beginning of each
chunk without subsequent intra-chunk updates.

\noindent{\textbf{The w/o tactile-prediction refinement variant.}}
The future-tactile prediction generated at the beginning of the chunk remains
fixed throughout execution. At each update offset, the latest tactile history
and the fixed chunk-start tactile latents are used to regenerate a full action
chunk, of which only the unexecuted suffix is applied.

\noindent{\textbf{The w/o future tactile prediction variant.}}
The future-tactile tokens and latent-alignment objective are removed, while
tactile-conditioned action updates are retained at the same intra-chunk
offsets.

\noindent{\textbf{The non-blocking joint refinement variant.}}
This variant changes only the execution schedule and uses the Full ReTouch
checkpoint.

\noindent{\textbf{The w/o Tactile-Patch Encoder variant.}}
The structured encoder is replaced with an MLP that encodes the flattened raw tactile signals before fusion, removing explicit finger-wise and patch-level spatial structure while keeping the remaining model architecture unchanged.

Except for the non-blocking joint refinement variant, all ablations are
trained separately.

\subsection{Diagnostic Comparison of Tactile-Patch Encoder 
Pretraining Configurations}
\label{app:tpe_pretraining_ablation}
\begin{table*}[t]
\centering

{\small
\setlength{\tabcolsep}{5pt}
\renewcommand{\arraystretch}{1.15}

\begin{tabular*}{\textwidth}
{@{\extracolsep{\fill}}llcc@{}}
\toprule
\textbf{Category}
& \textbf{Metric}
& \shortstack{\textbf{Single-head}\\\textbf{(Force only)}}
& \shortstack{\textbf{Balanced three-head}\\
\textbf{(Contact + Force + Distribution)}} \\
\midrule

Contact prediction
& Contact F1 $\uparrow$
& 0.9980
& \textbf{0.9999} \\
\midrule

Patch distribution
& Canonical patch-distribution KL $\downarrow$
& 0.1648
& \textbf{0.0048} \\
\midrule

\multirow{4}{*}{\shortstack[l]{Patch force\\reconstruction}}
& Active-force vector L2 $\downarrow$
& 2.1939
& \textbf{0.8801} \\

& Active-force direction cosine $\uparrow$
& 0.4089
& \textbf{0.8071} \\

& Inactive-force magnitude $\downarrow$
& \textbf{0.0094}
& 0.0260 \\

& Force-strength Pearson correlation $\uparrow$
& 0.7878
& \textbf{0.8643} \\

\bottomrule
\end{tabular*}
}
\caption{Diagnostic comparison of Tactile-Patch Encoder pretraining
configurations on frames sampled from the held-out offline test set. For the
single-head configuration, contact and distribution metrics are derived from
the predicted patch mean force. Force errors are reported in normalized
units.}
\label{tab:tactile_encoder_head_comparison}

\end{table*}
We compare a single-head force-reconstruction configuration with the balanced
three-head configuration used to initialize Full ReTouch. Both configurations
use the same encoder architecture, data split, random seed, and 20k-update
training budget. The single-head configuration predicts only the 3D patch
mean force, from which contact and patch-distribution predictions are derived.
The balanced three-head configuration directly predicts patch contact, the
canonical patch distribution, and contact-gated 3D patch mean forces, with
active- and inactive-patch force losses normalized separately. As shown in Table~\ref{tab:tactile_encoder_head_comparison}, the balanced
three-head configuration improves contact F1, canonical patch-distribution KL,
active-force vector L2, active-force direction cosine, and force-strength
correlation. The single-head configuration produces a lower inactive-force
magnitude. Overall, the balanced objective more accurately captures contact
distribution and active-contact forces. This diagnostic evaluates encoder pretraining quality and does not
by itself establish downstream manipulation gains.

\subsection{Effect of Recursive Tactile-Latent Refinement on Latent Alignment and Action Accuracy}
\label{app:recursive_refinement_analysis}

\begin{table*}[t]
\centering

{\small
\setlength{\tabcolsep}{5pt}
\renewcommand{\arraystretch}{1.08}

\begin{tabular*}{\textwidth}
{@{\extracolsep{\fill}}lcccccc}
\toprule
\textbf{Metric}
& \textbf{One-shot}
& \textbf{ReTouch}
& \textbf{$\Delta_{\mathrm{avg}}$}
& \textbf{$\Delta_4$}
& \textbf{$\Delta_8$}
& \textbf{$\Delta_{12}$} \\
\midrule

\multicolumn{7}{@{}l}{\textit{Pre-contact}} \\

Latent cosine $\uparrow$
& \textbf{0.98542}
& 0.98506
& $-0.037\%$
& $+0.011\%$
& $+0.023\%$
& $-0.149\%$ \\

Action-suffix MSE $\downarrow$
& 0.01435
& \textbf{0.01433}
& $+0.134\%$
& $+0.196\%$
& $-0.108\%$
& $+0.327\%$ \\

Arm-suffix MSE $\downarrow$
& \textbf{0.00221}
& 0.00222
& $-0.427\%$
& $-0.133\%$
& $-0.543\%$
& $-0.554\%$ \\

Hand-suffix MSE $\downarrow$
& 0.02042
& \textbf{0.02039}
& $+0.165\%$
& $+0.211\%$
& $-0.085\%$
& $+0.385\%$ \\

\midrule

\multicolumn{7}{@{}l}{\textit{In-contact}} \\

Latent cosine $\uparrow$
& 0.93879
& \textbf{0.94732}
& \textbf{$+0.908\%$}
& $+0.693\%$
& \textbf{$+1.148\%$}
& $+0.880\%$ \\

Action-suffix MSE $\downarrow$
& 0.02100
& \textbf{0.02055}
& \textbf{$+2.139\%$}
& $+1.317\%$
& $+2.074\%$
& \textbf{$+2.764\%$} \\

Arm-suffix MSE $\downarrow$
& \textbf{0.00339}
& 0.00340
& $-0.196\%$
& $-0.099\%$
& $-0.243\%$
& $-0.221\%$ \\

Hand-suffix MSE $\downarrow$
& 0.02981
& \textbf{0.02913}
& \textbf{$+2.272\%$}
& $+1.395\%$
& $+2.205\%$
& \textbf{$+2.937\%$} \\

\bottomrule
\end{tabular*}
}

\caption{Sample-matched offline diagnostics on the same held-out offline test set. At each update offset, both One-shot and ReTouch use the latest tactile
history to re-infer the action chunk. One-shot keeps the future tactile
latents predicted at offset zero fixed, whereas ReTouch recursively refines
them. Positive $\Delta$ denotes change in the preferred metric direction.
The One-shot and ReTouch columns aggregate eligible samples across offsets
4, 8, and 12, and $\Delta_{\mathrm{avg}}$ is computed from these aggregate
values. The columns $\Delta_4$, $\Delta_8$, and $\Delta_{12}$ report the
corresponding per-offset relative changes. Action errors are computed in
denormalized joint coordinates.}

\label{tab:recursive_refinement_diagnostics}
\end{table*}

We conduct sample-matched offline diagnostics comparing One-shot with ReTouch.
At offsets 4, 8, and 12, both methods use the latest tactile history to
re-infer the action chunk. One-shot keeps the future tactile latents predicted
at the beginning of the chunk fixed throughout execution, whereas ReTouch
recursively refines these latents at each update offset. The comparison
therefore isolates the effect of recursive tactile-latent refinement while
retaining tactile-conditioned action re-inference in both methods.

Contact onset is defined as the first stored frame at which the
$L_2$ magnitude of any raw three-axis taxel vector exceeds
$\theta_{\mathrm{onset}}=1.0$ in dataset-native units. No temporal persistence
criterion is applied. Samples before the onset frame are classified as
pre-contact, whereas samples at and after onset are classified as in-contact.

We evaluate update offsets $o\in\{4,8,12\}$. For each offset and contact
phase, let $N_o$ denote the number of eligible samples, and let
$\mathcal K_o$ contain the future tactile token indices whose represented
action intervals begin at or after $o$. The contact-phase index is omitted
from the notation for brevity. For comparison mode
$m\in\{\mathrm{one},\mathrm{ret}\}$, let
$\widehat{\mathbf z}^{m,(n)}_{o,k}$ denote the projected FAE state and let
$\mathbf z^{\mathrm{hid},(n)}_{o,k}$ denote the corresponding hindsight target.
We compute the token-wise latent cosine similarity as
\begin{equation}
S_{\mathrm{lat}}^{m}(o)=
\frac{1}{N_o|\mathcal K_o|}
\sum_{n=1}^{N_o}\sum_{k\in\mathcal K_o}
\cos\!\left(
\widehat{\mathbf z}^{m,(n)}_{o,k},
\mathbf z^{\mathrm{hid},(n)}_{o,k}
\right).
\label{eq:latent_metric}
\end{equation}
The One-shot mode reuses the offset-zero future tactile latents, whereas
ReTouch uses the recursively refined latents available at offset $o$.

For the action analysis, let $\widehat A^{m,(n)}_{o:}$ denote the predicted
denormalized joint-coordinate suffix from offset $o$, and let
$A^{(n)}_{o:}$ denote the corresponding ground-truth suffix. We compute
\begin{equation}
E_{\mathrm{act}}^{m}(o)=
\frac{1}{N_o}
\sum_{n=1}^{N_o}
\frac{
\left\|
\widehat A^{m,(n)}_{o:}-A^{(n)}_{o:}
\right\|_2^2
}{
18(16-o)
}.
\label{eq:action_metric}
\end{equation}
The denominator averages over the 18 joint coordinates and the remaining
$16-o$ action steps. At every update offset, both methods use the latest
tactile history to regenerate the action chunk. One-shot conditions action
re-inference on the fixed chunk-start tactile latents, whereas ReTouch
conditions it on the recursively refined tactile latents. Arm and hand errors
are additionally reported over their respective 6 and 12 joint coordinates.

All values are descriptive, sample-weighted aggregates. For latent cosine, a
positive relative change indicates increased similarity; for action MSE, a
positive relative change indicates reduced error. The One-shot and ReTouch
columns aggregate eligible samples across offsets 4, 8, and 12. The reported
$\Delta_{\mathrm{avg}}$ is computed from these aggregate values in the
preferred metric direction, whereas $\Delta_4$, $\Delta_8$, and
$\Delta_{12}$ denote the corresponding per-offset relative changes.

As shown in Table~\ref{tab:recursive_refinement_diagnostics}, neither metric
exhibits a consistent change across update offsets for pre-contact samples.
For in-contact samples, recursive tactile-latent refinement increases latent
similarity by $0.908\%$ and reduces action-suffix MSE by $2.139\%$ relative
to One-shot.


\end{document}